\documentclass[10pt,journal,cspaper,compsoc]{IEEEtran}

\ifCLASSOPTIONcompsoc
  \usepackage[nocompress]{cite}
\else
  \usepackage{cite}
\fi

\ifCLASSINFOpdf
   \usepackage[pdftex]{graphicx}
   \graphicspath{{../pdf/}{../jpeg/}}
   \DeclareGraphicsExtensions{.pdf,.jpeg,.png,.jpg}
\else
   \usepackage[dvips]{graphicx}
   \graphicspath{{../eps/}}
   \DeclareGraphicsExtensions{.eps}
\fi

\usepackage{amsmath}
\usepackage{amsfonts}

\ifCLASSOPTIONcompsoc
  \usepackage[caption=false,font=footnotesize,labelfont=sf,textfont=sf]{subfig}
\else
  \usepackage[caption=false,font=footnotesize]{subfig}
\fi

\usepackage{stfloats}
\usepackage{amsmath}
\usepackage{amsfonts}
\usepackage{enumerate}

\def\etal{{\em et al.}}

\newcommand{\myPara}[1]{\vspace{.05in}\noindent\textbf{#1}}

\usepackage{ragged2e}
\usepackage{booktabs}
\usepackage{color}
\usepackage{multirow}
\usepackage{bm}
\usepackage{bbm}
\usepackage{xcolor}
\usepackage{balance}
\usepackage{amssymb}
\usepackage[noend]{algpseudocode}
\usepackage{algorithmicx,algorithm}
\graphicspath{{./figures/}}
\usepackage{orcidlink}
\usepackage{booktabs}
\usepackage{multirow, graphicx}
\usepackage{makecell}
\makeatletter

\newcommand{\Rmnum}[1]{\expandafter\@slowromancap\romannumeral #1@}
\usepackage{pifont} 
\usepackage{bbding}     
\usepackage{fontawesome}
\usepackage{threeparttable}
\usepackage{float}
\makeatother

\begin{document}
\title{E\textsuperscript{3}NeRF: Efficient Event-Enhanced Neural Radiance Fields from Blurry Images}

\author{
Yunshan Qi$^{\orcidlink{0009-0004-8196-0085}}$,
Jia Li$^{\orcidlink{0000-0002-4346-8696}}$,~\IEEEmembership{Senior Member,~IEEE},
Yifan Zhao$^{\orcidlink{0000-0002-5691-013X}}$,~\IEEEmembership{Member,~IEEE},
Yu Zhang$^{\orcidlink{0000-0002-9653-3906}}$,~\IEEEmembership{Member,~IEEE},
Lin Zhu$^{\orcidlink{0000-0001-6487-0441}}$,~\IEEEmembership{Member,~IEEE}

\thanks{Yunshan Qi, Yifan Zhao, and Jia Li are with the State Key Laboratory of Virtual Reality Technology and Systems, School of Computer Science and Engineering, Beihang University, Beijing 100191, China.
E-mail: qi\_yunshan@buaa.edu.cn, zhaoyf@buaa.edu.cn, jiali@buaa.edu.cn}
\thanks{Lin Zhu is with the School of Computer Science, Beijing Institute of Technology, Beijing 100081, China.
E-mail: linzhu@bit.edu.cn}
\thanks{Yu Zhang is with the SenseTime Research and Tetras.AI.
E-mail: zhangyulb@gmail.com}
\thanks{Correspondence should be addressed to Jia Li and Lin Zhu. Website: http://cvteam.buaa.edu.cn}
\thanks{Our code, datasets, and detailed experimental results are publicly available on the project page: https://icvteam.github.io/E3NeRF.html}
}

\markboth{Submission to IEEE TRANSACTIONS ON PATTERN ANALYSIS AND MACHINE INTELLIGENCE}{Qi\MakeLowercase{\textit{et al.}}: E\textsuperscript{3}NeRF: Efficient Event-Enhanced Neural Radiance Fields from Blurry Images}

\IEEEtitleabstractindextext{%
\begin{abstract}
\justifying{
Neural Radiance Fields (NeRF) achieves impressive novel view rendering performance by learning implicit 3D representation from sparse view images.
However, it is difficult to reconstruct a sharp NeRF from blurry input \textcolor{black}{that} often occurs in the wild.
To solve this problem, we propose a novel Efficient Event-Enhanced NeRF (E\textsuperscript{3}NeRF), reconstructing sharp NeRF by utilizing both blurry images and corresponding event streams.
A blur rendering loss and an event rendering loss are introduced, which guide the NeRF training via modeling the physical image motion blur process and event generation process, respectively.
To improve the efficiency of the framework, we further leverage the latent spatial-temporal blur information in the event stream to evenly distribute training over temporal blur and focus training on spatial blur.
Moreover, a camera pose estimation framework for real-world data is built with the guidance of the events, generalizing the method to more practical applications.
Compared to previous image-based and event-based NeRF works, our framework makes more profound use of the internal relationship between events and images.
Extensive experiments on both synthetic data and real-world data demonstrate that E\textsuperscript{3}NeRF can effectively learn a sharp NeRF from blurry images, especially for high-speed non-uniform motion and low-light scenes.
}
\end{abstract}

}

\maketitle
\IEEEdisplaynontitleabstractindextext
\IEEEpeerreviewmaketitle

\section{Introduction}
\label{sec:1}

\begin{figure*}[!t]
    \centering
    \includegraphics[width=1\linewidth]{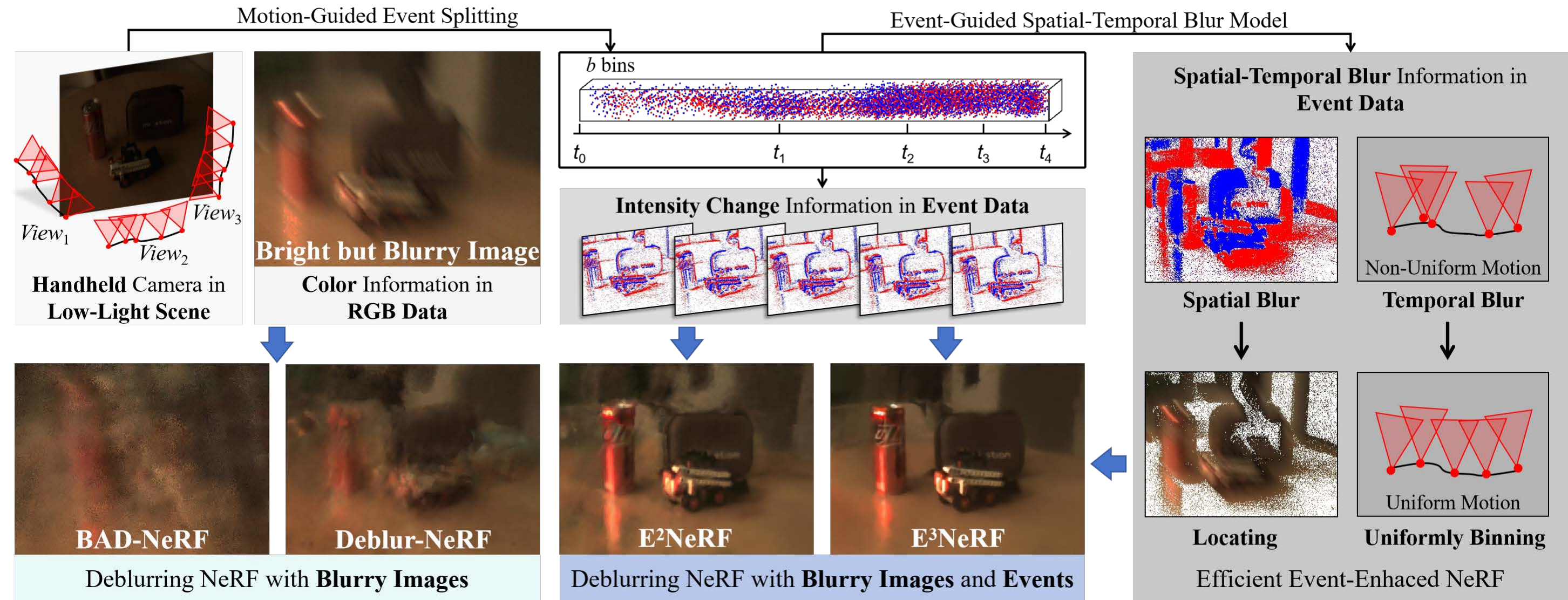}
    \caption{
    In a low-light static scene, handheld traditional cameras often capture blurry images.
    Image-based deblurring NeRF, such as BAD-NeRF\cite{bad-nerf} and Deblur-NeRF\cite{Deblur-NeRF}, will fail to synthesize sharp novel view images when facing severe blur.
    With an event camera, we can capture the event streams corresponding to the blurry images.
    By using the light intensity change information in event data, E\textsuperscript{2}NeRF\cite{e2nerf} achieves a primary deblurring effect.
    In E\textsuperscript{3}NeRF, we further extract and utilize the spatial-temporal blur information in events to spread training evenly on temporal blur and focus training on spatial blur.
    Additionally, we use motion-guided event splitting to determine the suitable number of event bins for each view.
    Eventually, E\textsuperscript{3}NeRF realizes impressive implicit 3D representation learning results and high efficiency training in complex scenes with severely blurry input.
    }
    \label{fig:1}
\end{figure*}

\IEEEPARstart{W}{ith} the advent of Neural Radiance Fields (NeRF)\cite{NeRF}, significant progress has been made in neural 3D representation and novel view synthesis tasks in the past few years.
NeRF takes 3D location and 2D \textcolor{black}{viewing} direction as input and uses multi-view images of objects or scenes as supervision to learn the neural volumetric representation, which is parameterized as a multilayer perceptron (MLP).
To generate high-fidelity novel view synthesis results, NeRF encodes the input position information into higher dimensions and utilizes volume rendering techniques for the output of the network (color and density) during both the training and testing processes.

The premise that NeRF can produce impressive results relies on the assumption that the input image quality is of high standards, devoid of blurs, and has sufficient lighting.
However, obtaining such high-quality images can be challenging in many real-world scenes.
As shown in the left part of Fig.~\ref{fig:1}, traditional cameras often capture blurry images due to handheld operation and long exposure time in low-light scenes, presenting significant challenges for original NeRF.
Some deblurring NeRF works, such as BAD-NeRF \cite{bad-nerf} and Deblur-NeRF \cite{Deblur-NeRF}, are tailored for this situation but encounter difficulties in managing substantial motion.
Furthermore, the initial pose generation of these methods is not robust to some extremely blurred scene images, and the time-based uniform camera pose interpolation in BAD-NeRF fundamentally contradicts the non-uniform camera motion characteristics inherent in the real-world blurring process.
Consequently, relying solely on blurry RGB images to reconstruct sharp NeRF proves to be a considerable obstacle when addressing such scenarios.

In contrast to relying solely on images, combining additional information to guide the neural radiance fields learning process is promising.
Event camera is a bio-inspired vision sensor that measures the brightness changes of each pixel asynchronously.
Unlike traditional frame-based cameras, event cameras can record high temporal resolution and high dynamic range visual information of the scene.
The high temporal resolution event stream makes up for the spatial-temporal information insufficiency of blurry images captured by traditional frame-based cameras.
Therefore, event-based image deblurring has become an attractive research topic in recent years \cite{jiang2020learning}, \cite{lin2020learning}, \cite{shang2021bringing}, \cite{xu2021motion}, \cite{pan2019bringing}.
Inspired by this, we introduce the event stream into the learning process of neural radiance fields and propose an \textbf{E}fficient \textbf{E}vent-\textbf{E}nhanced framework (E\textsuperscript{3}NeRF) to solve the reconstruction quality degradation caused by blurry input images.

This paper aims to investigate ``how to derive a sharp NeRF from blurry images induced and corresponding events caused by non-uniform intense camera motion in the context of low-light scenes''.
Our insight is to explore the utilization of spatial-temporal blur information and light change information in the asynchronously acquired high temporal resolution event streams to enhance the learning of NeRF.
As shown in the right part of Fig.~\ref{fig:1}, we leverage the spatial-temporal distribution of the events to guide the learning evenly over temporal blur and focus the learning on spatial blur areas.
We also utilize the light change information in events to enhance the sharp NeRF reconstruction from blurry images, as in the middle part of Fig.~\ref{fig:1}.
Specifically, we first discretely predict the latent sharp images along with the camera motion trajectory during exposure time for each view through the NeRF network.
The predicted blurry images are obtained by taking the weighted average of the predicted sharp images and are compared with the input blurry images to calculate the blur rendering loss.
The predicted events are simulated from the predicted sharp images and compared with the input events to obtain the event rendering loss, which refines the neural 3D representation learning.
We also design an event-guided pose estimation framework to obtain pose sequences for the virtual sharp images, ensuring our method remains effective for real-world scenes without ground-truth poses for referring.
Due to the augmentation of the network with event data, E\textsuperscript{3}NeRF can learn a sharp NeRF efficiently, achieving not only the deblurring effect on the input images but also high-quality novel view synthesis with severely blurry input images, especially in low-light scenes with high-speed non-uniform camera motion.

To the best of our knowledge, this is the first work to reconstruct a sharp NeRF using both event and RGB (ERGB) data.
Our contributions can be summarized as follows:

1) We propose an efficient event-enhanced neural radiance fields (E\textsuperscript{3}NeRF) framework for reconstructing a sharp NeRF from blurry images and corresponding events. 
Two novel losses are introduced to exploit the effect of high temporal resolution light change information in events for the sharp NeRF reconstruction with blurry input.

2) We further extract the latent spatial-temporal blur information from the event stream, which focuses training evenly on areas where blur occurs, improving training efficiency and robustness for high-speed and non-uniform camera motion.
An event-guided pose estimation framework is also designed for real-world scenes, significantly enhancing the practicability of our method.

3) We build both synthetic and real-world datasets to train and test our model. Experimental results demonstrate that our method outperforms existing methods. Additionally, we provide a benchmark for future research on NeRF reconstruction from blurry images and event streams.

A preliminary version of this work (E\textsuperscript{2}NeRF\cite{e2nerf}) has been published in ICCV 2023.
The main extensions include the incorporation of an event-guided spatial-temporal blur model grounded in temporal blur uniform binning, spatial blur locating, and a motion-guided event-splitting strategy.
These components are specifically designed to address scenarios with non-uniform camera motion blur and varying degrees of blur across different views, thereby improving the efficiency of the model.
Additionally, we construct extra synthetic and real-world datasets, encompassing more severely blurry images with non-uniform motion and ground-truth sharp images for evaluation. 
Numerous experimental analyses are also conducted on the new synthetic and real-world datasets to showcase the satisfactory performance of our model.
A recent work, Ev-DeblurNeRF \cite{evdeblurnerf}, shares a similar goal to ours, but uses continuous dense view events and requires the corresponding ground-truth poses as reference for training in real-world scenes as discussed in Sec.~\ref{sec:2.3}. 

The rest of this paper is organized as follows.
Sec.~\ref{sec:2} reviews the related works of NeRF, image deblurring, event camera, and event-related NeRF.
Sec.~\ref{sec:3} and Sec.~\ref{sec:4} introduce the background and framework of E\textsuperscript{3}NeRF, separately.
Sec.~\ref{sec:5} presents the proposed datasets and public datasets used in this paper. The compared methods, experimental settings, as well as the discussions and analysis, are also included in this section.
Finally, the paper is concluded in Sec.~\ref{sec:6}.

\begin{table*}[t]
    \caption{
    A comparison of previous deblurring NeRF works, event-based NeRF works, ERGB-based deblurring NeRF works, and our methods.}
    \label{tab:1}
    \centering
    \scriptsize
    \begin{tabular}{c||c|c|c|c}
        \toprule 
        
        Method & Input Images  & Input Events & Poses Estimation for Real-World Scenes & Objective \\
                    
        \midrule
        NeRF \cite{NeRF}      & Sparse View \textbf{\emph{Sharp}} Images & N/A & SfM with \textbf{\emph{Sharp}} Images
        & NeRF \\
        \midrule
        Deblur-NeRF \cite{Deblur-NeRF}     & \multirow{3}*{\makecell[c]{Sparse View \textbf{\emph{Blurry}} Images \\ (Forward-Facing)}} & \multirow{3}*{N/A} & \multirow{3}*{SfM with \textbf{\emph{Blurry}} Images}
        & \multirow{3}*{Deblurring NeRF} \\
        BAD-NeRF \cite{bad-nerf}        & & & & \\
        DP-NeRF  \cite{dpnerf}        & & & & \\
        \midrule
        Ev-NeRF  \cite{enerf1}        
        & \multirow{4}*{N/A} & \multirow{4}*{\makecell[c]{\textcolor{black}{\textbf{\emph{Continuous Dense View}}} \\ \textcolor{black}{Event Stream}}} 
        & SfM with Continuous Image Sequence
        & \multirow{4}*{Event-Based NeRF} \\
        \cline{4-4}
        EventNeRF \cite{eventnerf}              & & & \multirow{3}*{\makecell[c]{N/A \\ (Need Given Ground-Truth Poses)}} & \\
        \emph{e}-NeRF  \cite{renerf}         & & & &  \\
        Deblur \emph{e}-NeRF  \cite{deblurenerf}  & & & &  \\
        
        \midrule
        E-NeRF \cite{enerf2}                  
        & \multirow{2}*{\makecell[c]{Sparse View \textbf{\emph{Blurry}} Images \\ (Forward-Facing)}} 
        & \multirow{2}*{\makecell[c]{\textcolor{black}{\textbf{\emph{Continuous Dense View}}} \\ \textcolor{black}{Event Stream}}}
        & \multirow{2}*{\makecell[c]{N/A \\ (Need Given Ground-Truth Poses)}} & ERGB-Based Deblurring NeRF \\
        
        Ev-DeblurNeRF \cite{evdeblurnerf}           & & & & \textcolor{black}{(\textbf{\emph{Dense View}} Events)} \\
        
        \midrule
        E\textsuperscript{2}NeRF (ours) 
        & \multirow{2}*{Sparse View \textbf{\emph{Blurry}} Images} 
        & \multirow{2}*{\makecell[c]{\textcolor{black}{\textbf{\emph{Discrete Sparse View }}} \\ \textcolor{black}{Event Streams} }}
        
        & \multirow{2}*{SfM with \textbf{\emph{Blurry}} Images and Events} & ERGB-Based Deblurring NeRF \\
        
        E\textsuperscript{3}NeRF (ours) & & & & \textcolor{black}{(\textbf{\emph{Sparse View}} Events)} \\
        
        \bottomrule
    \end{tabular}

    \begin{tablenotes}
        \scriptsize
        \item \textbf{SfM}: The structure-from-motion method COLMAP \cite{colmap}.
        \item \textbf{Forward-Facing}: Only effective for the input views that are from the front of the scene.
        \item \textcolor{black}{\textbf{Continuous Dense View Event Stream}: A continuous event stream captured by a camera continuously moving in the scene, as shown in the right part of Fig.~\ref{fig:2}.}
        \item \textcolor{black}{\textbf{Discrete Sparse View Event Streams}: The event streams captured during the exposure times of discrete sparse views images, as shown in the left part of Fig.~\ref{fig:2}.}
        
    \end{tablenotes}
\end{table*}

\begin{figure}[t]
    \centering
    \color{black}
    \includegraphics[width=1\linewidth]{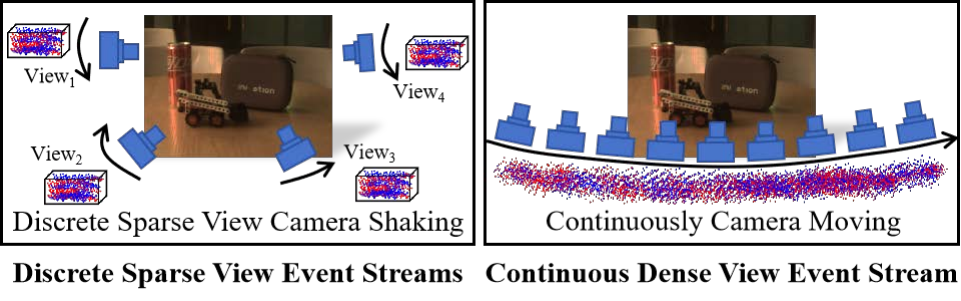}
    \caption{Comparison of input events under two different settings in Table~\ref{tab:1}.}
    \label{fig:2}
\end{figure}

\section{Related Work}
\label{sec:2}

\subsection{Neural Radiance Fields}
\label{sec:2.1}
In the past few years, NeRF\cite{NeRF} has achieved impressive results and attracted much attention for neural implicit 3D representation and novel view synthesis tasks.
FastNeRF\cite{fastnerf} and Depth-supervised NeRF\cite{depthnerf} improve the learning speed of NeRF.
Neural scene flow fields\cite{flownerf} explores 3D representation learning of dynamic scenes.
pixelNeRF\cite{pixelnerf} and RegNeRF\cite{regnerf} try to use a small number of input images to achieve high-quality novel view synthesis.
Mip-NeRF\cite{mipnerf} proposes a frustum-based sampling strategy to implement NeRF-based anti-aliasing, solving the artifacts problem and improving the training speed.
Some works aim to improve NeRF with low-quality input images like NeRF in the wild\cite{nerfinw}, NeRF in the dark\cite{nerfind}, and HDR-NeRF\cite{hdr-nerf}.
In the context of reconstructing a sharp NeRF from blurry images, Deblur-NeRF \cite{Deblur-NeRF} introduces the deformable sparse kernel to model the blurring process, enabling the achievement of sharp NeRF reconstruction from blurry images.
BAD-NeRF \cite{bad-nerf} jointly learns a sharp NeRF and recovers the camera motion trajectories during the exposure times.
DP-NeRF \cite{dpnerf} proposes a rigid blurring kernel and an adaptive weight proposal based on two physical priors, 3D consistency and the relationship between depth and blur, to constrain the training.
However, with only blurry images as input, the reconstruction results of these three methods are limited, and their pose initialization could fail with severely blurry input.
Additionally, due to the optimization of pose and blur kernel during training, they can only be effective on the forward-facing input as shown in the TABLE~\ref{tab:1}.

\subsection{Image Deblurring}
\label{sec:2.2}
A blurred image can be expressed as a sharp image multiplied by a blur kernel plus noise.
However, due to the non-uniqueness of the blur kernel, the deblurring problem becomes ill-posed.
Traditional algorithms use hand-crafted or sparse priors to predict the blur kernel\cite{sdeblur1},\cite{sdeblur2},\cite{sdeblur3}.
With the development of deep learning, some works attempt to directly learn an end-to-end mapping from blurry images to sharp images with neural networks \cite{sdeblur4}, \cite{sdeblur5}, \cite{sdeblur6}.
Tao~\etal{}\cite{vdeblur8} import the “coarse-to-fine” strategy into the deblurring network.
Zamir~\etal{}\cite{sdeblur7} introduce a novel per-pixel adaptive design to reweight the local features with encoder-decoder architectures.
Both works achieve state-of-the-art performance for single-image deblurring.
However, in real-world scenarios, the intricate and varied camera motion and the absence of light intensity change information during the exposure time of traditional images fundamentally limit the performance of the image deblurring methods.

\subsection{Event-Based Vision in Neural Radiance Fields}
\label{sec:2.3}
\myPara{Event Camera:}
Dynamic vision sensor (DVS)\cite{dvs}, also known as event camera, can asynchronously generate events when the brightness change of any pixel reaches the threshold, which effectively overcomes the problem of information loss during and between the exposure times of traditional cameras.
Dynamic active vision sensor (DAVIS)\cite{davis} realizes the simultaneous acquisition of RGB images and events, which attracts widespread attention in the computer vision community.
At present, event-based computer vision algorithms have achieved remarkable results in optical flow estimation\cite{eflow1,eflow2,eflow3,eflow4,eflow5}, depth estimation\cite{edepth1,edepth2,edepth3,edepth4}, feature detection and tracking\cite{etrack1,etrack2,etrack3} as well as simultaneous localization and mapping\cite{eslam1,eslam2,eslam3}.
In addition, to address the lack of event-based datasets, some event simulators \cite{esim1,esim2,esim3} are designed to simulate events through videos.
With the high temporal resolution, event data also has significant advantages in image deblurring.
For example, Pan~\etal{}\cite{pan2019bringing} propose an event-based double integral model and realize the image deblurring with events.
Shang~\etal{}\cite{shang2021bringing} develop D2Net for video deblurring with events.

\myPara{Event-Based NeRF:}
Recently, Ev-NeRF \cite{enerf1} aims to learn a grayscale NeRF from the event stream, and EventNeRF \cite{eventnerf} can further learn a color NeRF from color events.
\textit{e}-NeRF \cite{renerf} explores the robustness of event-based NeRF under different camera parameters and non-uniform camera motion.
Deblur \textit{e}-NeRF \cite{deblurenerf} improves \textit{e}-NeRF by solving the motion blur problem of events.
\textcolor{black}{
However, these works all need a continuous dense view event stream captured by a camera continuously moving around the scene as input, as shown in the right part of Fig.~\ref{fig:2}.
Accordingly, poses corresponding to the dense view events are also required for training, leading to the pose estimation being a challenging problem for real-world scenes.
For instance, Ev-NeRF generates the poses with a continuous dense image sequence corresponding to the events for reference, and other methods need ground-truth poses for training, as shown in TABLE~\ref{tab:1}.
Besides, the reconstruction results of these works always have artifacts and chromatic aberration due to the lack of absolute RGB values for supervision.}

\myPara{ERGB-Based NeRF:}
Some concurrent works attempt to reconstruct NeRF using the hybrid ERGB visual data for different objectives.
DE-NeRF \cite{denerf} aims to learn a dynamic deformable NeRF.
Be-NeRF \cite{benerf} trains a separate NeRF for each view, achieving video synthesis from a single blurry image and corresponding events, struggling in novel view synthesis on unseen views.
\textcolor{black}{
E-NeRF \cite{enerf2} proposes a continuous event-by-event loss for the NeRF training.
Ev-DeblurNeRF \cite{evdeblurnerf} further exploits the use of this loss for deblurring NeRF based on the DP-NeRF framework, which limits it to forward-facing scenes.
Besides, like the event-based NeRF works, E-NeRF and Ev-DeblurNeRF both require a continuous dense view event stream that not only corresponds to the sparse view blurry images, but also connects the adjacent image views.
Therefore, they also need a large number of events and ground-truth poses for training, as shown in TABLE~\ref{tab:1}.
}

Unlike prior works, our E\textsuperscript{3}NeRF imposes no additional restrictions on input views, images, events, and camera poses. It emphasizes event representation in blurry images and encompasses a novel blur-solving method based on the spatial-temporal distribution and high temporal resolution light-intensity-change measuring characteristic of event stream, yielding superior results and demonstrating robust generalization across real-world complex scenes, particularly in cases of non-uniform camera motion.

\section{Background}
\label{sec:3}

\subsection{Neural Radiance Fields}
\label{sec:3.1}
NeRF takes sparse views images as input to reconstruct 3D scenes and synthesize novel view images.
The core of NeRF\cite{NeRF} is to learn 3D volume representation via a multilayer perceptron (MLP).
The input is 3D position $\mathbf{o}$ and 2D observed ray direction $\mathbf{d}$ of a point in the scene, and the output is the color $\mathbf{c}$ and density $\sigma$ of the point:
\begin{equation}
    (\mathbf{c},\sigma) = F_{\mathbf{\theta}}(\gamma_{o}(\mathbf{o}),\gamma_{d}(\mathbf{d})),
    \label{eq:1}
\end{equation}
where $F_{\mathbf{\theta}}$ is the MLP network with parameters $\mathbf{\theta}$.
$\gamma_{o}(\cdot)$ and $\gamma_{d}(\cdot)$ in Eq.~\eqref{eq:1} are defined as:
\begin{equation}
    \gamma_{M}(x) = \{\sin(2^{m}\pi{x}), \cos(2^{m}\pi{x})\}_{m=0}^{M},
    \label{eq:2}
\end{equation}
which maps the input 5D coordinates into a higher-dimensional space, enabling the network to better learn the color and geometry information of the scene.
We set $M=10$ for position $\mathbf{o}$ and $M=4$ for direction $\mathbf{d}$ as in NeRF.

To render images of different views from the implicit 3D scene representation, NeRF uses the classical volume rendering method.
For a given ray $\mathbf{r}(l)=\mathbf{o}+l\mathbf{d}$ emitting from camera center $\mathbf{o}$ with direction $\mathbf{d}$, its expected color projected on the pixel $\mathbf{x}(x_p,y_p)$ is:
\begin{equation}
    \begin{aligned}
        \hat{C}(\mathbf{r},\mathbf{x}) = \sum_{i=1}^{N}T_{i}(1-\exp(-{\sigma}_i{\delta}_i))\mathbf{c}_i,\\
        \text{where} \quad T_i = \exp(-\sum_{j=1}^{i-1}\sigma_{j}\delta_{j}).
    \end{aligned}
    \label{eq:3}
\end{equation}
The ray is divided into $N$ discrete bins, with $l_{n}$ and $l_{f}$ representing the near and far bounds, respectively.
$\mathbf{c}_{i}$ and $\sigma_{i}$ are the output of $F_{\mathbf{\theta}}$, indicating the color and density of each bin through which the ray passes.
$\delta_{i}=l_{i+1}-l_{i}$ is the distance between adjacent sampled bins.
$T_{i}$ is the transparency of the particles between $l_{n}$ and sampled $i$-th bin.
We can also obtain the depth information of the scene for ray $\mathbf{r}$ on pixel $\mathbf{x}$ by modifying Eq.~\eqref{eq:3} as:
\begin{equation}
    \begin{aligned}
        D(\mathbf{r},\mathbf{x}) = \sum_{i=1}^{N}T_{i}(1-\exp(-{\sigma}_i{\delta}_i))l_i,
    \end{aligned}
    \label{eq:4}
\end{equation}
where $\{l_i\}_{i=1}^{N}$ can be regarded as depth of the $i$-th bin.

The final loss of NeRF is the sum of mean squared losses between the predicted color $\hat{C}(\mathbf{x})$ and input image color $C(\mathbf{x})$ for all pixels $\mathbf{x} \in \mathbf{X}_{v}$ of each view.
Note that we omit $\mathbf{r}$ in $\hat{C}(\mathbf{r}, \mathbf{x})$ here because $\mathbf{r}$ and $\mathbf{x}$ have a one-to-one correspondence at the same view.
To achieve reasonable sampling for the model, NeRF uses the hierarchical volume sampling strategy, which optimizes the coarse model $\hat{C}_{c}(\mathbf{x})$ and fine model $\hat{C}_{f}(\mathbf{x})$ simultaneously:
\begin{equation}
    \mathcal{L}= \sum_{v=1}^{V}\sum_{\mathbf{x}\in\mathbf{X}_{v}}[\|\hat{C}_{c}(\mathbf{x})-C(\mathbf{x})\|_{2}^{2}+\\
    \|\hat{C}_{f}(\mathbf{x})-C(\mathbf{x})\|_{2}^{2}],
    \label{eq:5}
\end{equation}
where $V$ is the number of input views.
The density obtained by the coarse network is applied to determine the sampling weight of the fine network.

\subsection{Event Generation}
\label{sec:3.2}
Event camera generates an event $\mathbf{e}(x,y,\tau,p)$ asynchronously when the changes of the brightness at pixel $(x,y)$ in the log domain reach the threshold $\Theta$ at time $\tau$:
\begin{equation}
    p_{x,y,\tau} = 
    \begin{cases}
    -1,&\log(\mathcal{I}_{x,y,\tau})-\log(\mathcal{I}_{x,y,\tau-\Delta\tau})< -\Theta\\
    +1,&\log(\mathcal{I}_{x,y,\tau})-\log(\mathcal{I}_{x,y,\tau-\Delta\tau})>\Theta
    \end{cases},
    \label{eq:6}
\end{equation}
where $p$ indicates the direction of brightness change, $\mathcal{I}_{(x,y,\tau)}$ is the brightness value of pixel $(x,y)$ at time $\tau$.
In E\textsuperscript{2}NeRF \cite{e2nerf}, we divide an given event stream, $\{\mathbf{e}_{i}\}_{t_{start}<\tau_{i}\leq t_{end}}$ corresponding to a blurry image with exposure time $t_{exp}=t_{end}-t_{start}$, into $b$ event bins equally by time:
\begin{equation}
    \{B_{k}'\}_{k=1}^{b} = \{\mathbf{e}_{i}(x_{i},y_{i},\tau_{i},p_{i})\}_{t_{k-1}<\tau_{i}\leq t_{k}}.
    \label{eq:7}
\end{equation}
$t_{k}=t_{start}+\frac{k}{b}t_{exp}$ is the time division points of the bins.

\section{Method}
\label{sec:4}
In our settings, E\textsuperscript{3}NeRF takes blurry images and the corresponding events from $V$ sparse views as input to reconstruct a sharp NeRF and synthesize sharp novel view images.
Since the operations for all views are the same, we use one view as an example to illustrate our algorithms in this section.
In Sec.~\ref{sec:4.1} we clarify the correlation between image motion blur and events, introducing an event-guided spatial-temporal blur model to improve the performance and efficiency of E\textsuperscript{3}NeRF.
In Sec.~\ref{sec:4.2} we present the blur rendering loss and event rendering loss to introduce events into the NeRF training process.
In Sec.~\ref{sec:4.3} we formulate an event-guided pose estimation framework for real-world data based on the input events and blurry images, enhancing the practical applicability of E\textsuperscript{3}NeRF in real scenarios.

\subsection{Event-Guided Spatial-Temporal Blur Model}
\label{sec:4.1}

\subsubsection{Correlation between Image Motion Blur and Events}
\label{sec:4.1.1}

Traditional cameras convert the number of photons hitting the sensor during the exposure time into voltage values to record the lightning information.
According to this, the physical formation of an image $I$ can be expressed as the integration of consecutive virtual sharp images $I_{vir}(t)$ with normalization factor $\phi$:
\begin{equation}
    \begin{aligned}
        {I} = \phi\int_{t_{start}}^{t_{end}}I_{vir}(t)\mathrm{d}t,\\
    \end{aligned}
    \label{eq:8}
\end{equation}
where $t_{start}$ and $t_{end}$ are the times when the exposure starts and ends.
An motion-blurred image $I_{blur}$ is caused by the change of $I_{vir}(t)$ during exposure time on pixel $\mathbf{x}=(x,y)$:
\begin{equation}
    \begin{aligned}
        \{\exists\ (\mathbf{x},\tau-\Delta \tau, \tau)| 
        & I_{vir}(\tau-\Delta \tau,\mathbf{x}) \neq I_{vir}(\tau,\mathbf{x}), \\
        & t_{start} \leq \tau-\Delta \tau < \tau \leq t_{end}\}.
    \end{aligned}
    \label{eq:9}
\end{equation}
$\mathbf{x}$ denotes where blur occurs (spatial blur) and $\tau$ denotes when blur occurs (temporal blur).
Correspondingly, if the change reaches the threshold of the event camera, the event $\mathbf{e}(x,y,\tau,p)$ in Eq.~\eqref{eq:6} is triggered and locates the above-mentioned pixel with $(x,y)$ and time with $\tau$ respectively.

At this point, we establish a correlation between spatial-temporal image motion blur and events: the events corresponding to a blurry image indicate where and when blur occurs.
According to this, we use the spatial-temporal distribution of events to guide training evenly distributed on temporal blur as in Sec.~\ref{sec:4.1.2} and focus training on spatial blur as in Sec.~\ref{sec:4.1.3}.

\begin{figure}[t]
    \centering
    \includegraphics[width=1\linewidth]{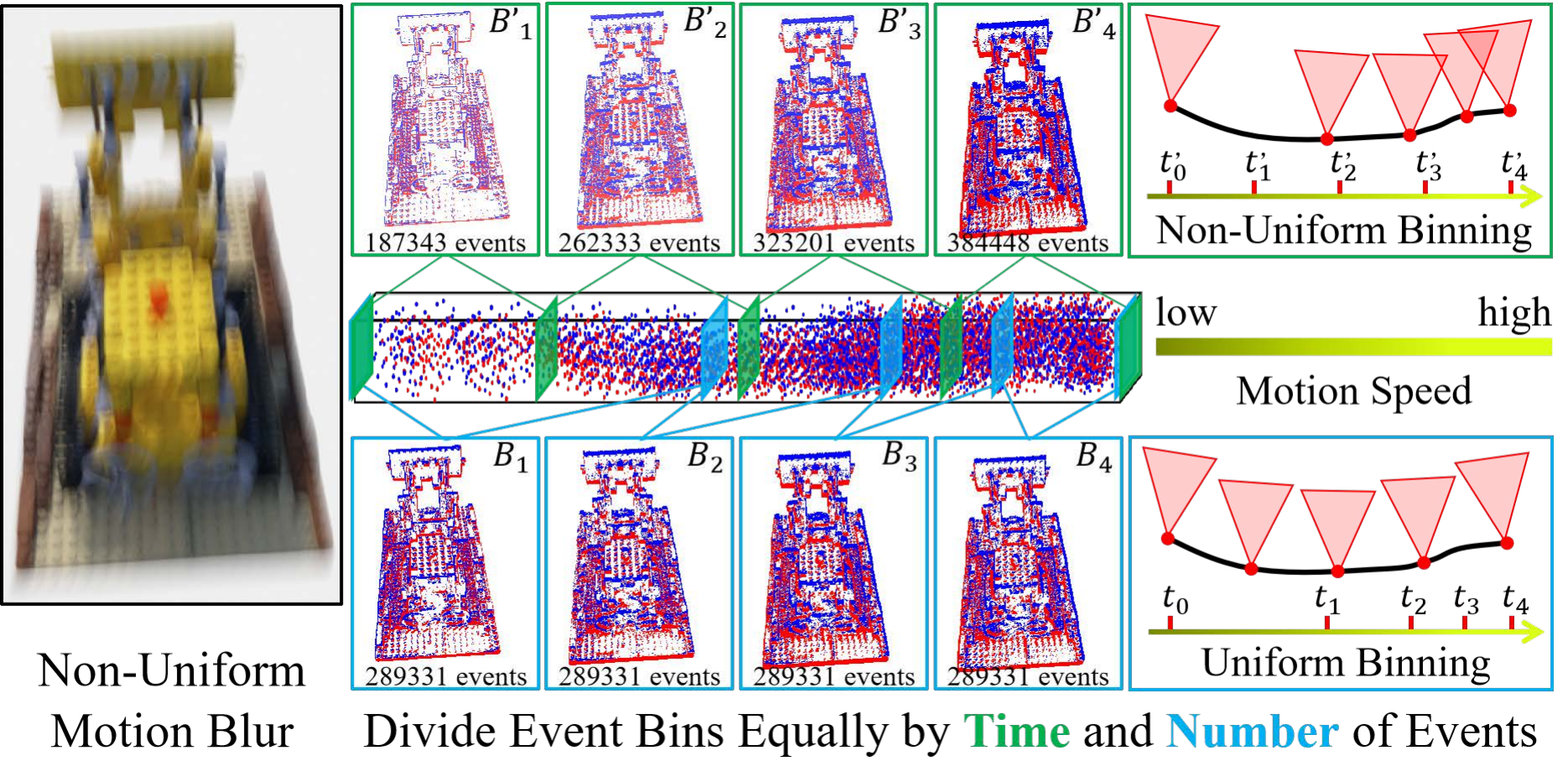}
    \caption{
    Temporal blur uniform binning. The left part of the figure is a blurry image caused by non-uniform motion, and the middle part is the corresponding event stream and visualized event bins. The right part of the figure shows the camera motion blur trajectory, motion speed, and the corresponding time axis. With the event-guided temporal blur binning, the time division points of event bins are concentrated on the moments with high motion speed (yellow arrow), and the camera poses corresponding to the time points are evenly distributed on the camera motion trajectory (red triangles on the black curve in the blue box).
    }
    \label{fig:3}
\end{figure}

\subsubsection{Event-Guided Temporal Blur Uniform \textcolor{black}{Binning}}
\label{sec:4.1.2}

As shown in Fig.~\ref{fig:3}, the camera motion during exposure time is always non-uniform in speed and the density of the event stream varies with the changing of speed, indicating that the blur degree varies at different moments.
In E\textsuperscript{2}NeRF, we divide event bins equally by time with Eq.~\eqref{eq:7}, ignoring the temporal blur distribution of the blurry image in the events data.
As in the green boxes, the numbers of events in different bins are unbalanced, and the temporal motion blur is non-uniformly binned, which will lead to imbalanced training and performance degradation.
Therefore, considering the temporal distribution of events, we divide $\{\mathbf{e}_{i}\}_{t_{start}<\tau_{i}\leq t_{end}} = \{\mathbf{e}_{i}\}_{i=1}^s$ corresponding to a blurry image into $b$ event bins equally by the number of events:
\begin{equation}
    \{B_{k}\}_{k=1}^{b} = \{\mathbf{e}_{i}(x_{i},y_{i},\tau_{i},p_{i})\}_{ \lfloor \frac{s(k-1)}{b} \rfloor <i\leq \lfloor \frac{sk}{b} \rfloor},
    \label{eq:10}
\end{equation}
where $s$ is the number of the events during the exposure time and $\{\mathbf{e}_{i}\}_{i=1}^s$ is sorted by the timestamps $\{\tau_i\}_{i=1}^s$.
Then we can get $b+1$ time division points of the event bins:
\begin{equation}
    t_{0}=t_{start}, t_{b}=t_{end}, \{t_{k}\}_{k=1}^{b-1} = \tau_{ \lfloor \frac{sk}{b} \rfloor}, 
    \label{eq:11}
\end{equation}
which uniformly divides the temporal blur.
Consequently, the time division points are concentrated on the moments with high motion speed, and the corresponding camera poses are evenly distributed on the camera motion trajectory as shown in the right part of Fig.~\ref{fig:3}.

The event-guided temporal blur uniform binning using ``when the events are triggered'' evenly distributes $\mathcal{L}_{event}$ in Sec.~\ref{sec:4.2.2} on the camera motion trajectory, stabilizing network training and generating better results, especially for non-uniform motion blur.

\subsubsection{Event-Guided Spatial Blur Locating}
\label{sec:4.1.3}

\begin{figure*}[t]
    \centering
    \includegraphics[width=1\linewidth]{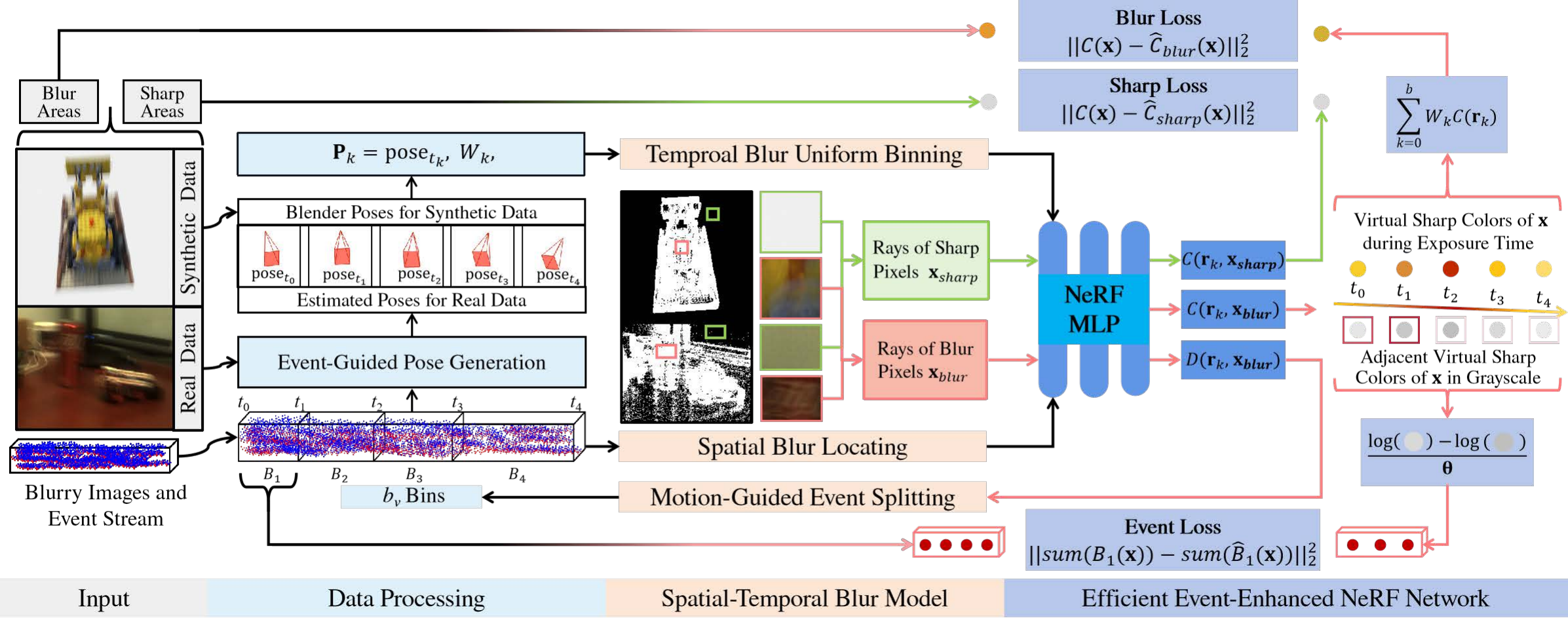}
    \caption{
    The architecture of E\textsuperscript{3}NeRF.
    The input is sparse views of blurry images and corresponding event streams.
    The figure shows the operation of one view as an example.
    For real-world data, we use the event-guided pose estimation model to obtain the pose sequence.
    For synthetic data, we use ground-truth poses as in NeRF.
    Then, we use the event-guided spatial-temporal blur model to focus the training on areas with spatial blur and distribute the training evenly over temporal blur.
    Simultaneously, we use motion-guided event splitting to split events for each view individually.
    For blurry pixels, as shown with the red arrows, the network renders $b+1$ virtual sharp colors, with which we calculate the predicted blurry color $\hat{C}_{blur}(\mathbf{x})$ and event bin $\hat{B}_{k}(\mathbf{x})$.
    Then, comparing with input color $C(\mathbf{x})$ and event bin $B_{k}(\mathbf{x})$, we get the proposed blur loss and event loss. 
    For sharp pixels, as shown with the green arrows, we solely conduct a sharp loss as in NeRF.
    The above operation is repeated for each view during training.
    }
    \label{fig:4}
\end{figure*}

As shown in the middle of Fig.~\ref{fig:4}, not all areas of the input image are affected by the motion blur, because camera motion does not necessarily result in pixel brightness changes on smooth areas (green bounding box in Fig.~\ref{fig:4}) and the areas with texture details are more likely to produce pixel brightness changes and trigger Eq.~\eqref{eq:9} with camera motion, resulting in motion blur (red bounding box in Fig.~\ref{fig:4}).
Therefore, according to the spatial distribution of events, we can categorize pixels of a blurry image into blurred texture detailed areas and sharp smooth areas:
\begin{equation}
    \begin{aligned}
        \mathbf{X}_{blur}=\{\mathbf{x}| \exists (\tau, p), \mathbf{e}(\mathbf{x}, \tau, p) \in \{\mathbf{e}_i\}_{i=1}^s \},\\
        \mathbf{X}_{sharp}=\{\mathbf{x}| \forall (\tau, p), \mathbf{e}(\mathbf{x}, \tau, p) \notin \{\mathbf{e}_i\}_{i=1}^s \}.
    \end{aligned}
    \label{eq:12}
\end{equation}
Compared to treating all pixels of the input image as blurred in E\textsuperscript{2}NeRF, Eq.~\eqref{eq:12} can guide where to use the computationally expensive operations for blur areas and where to use the computationally cheap operations for sharp areas, which are shown in Fig.~\ref{fig:4} as the red and green arrows, respectively.

The event-guided spatial blur locating using ``where the events are triggered'' focuses the training on spatial blur and directs computing resources in areas with texture details, improving the network efficiency and performance.

\begin{figure}[t]
    \centering
    \includegraphics[width=1\linewidth]{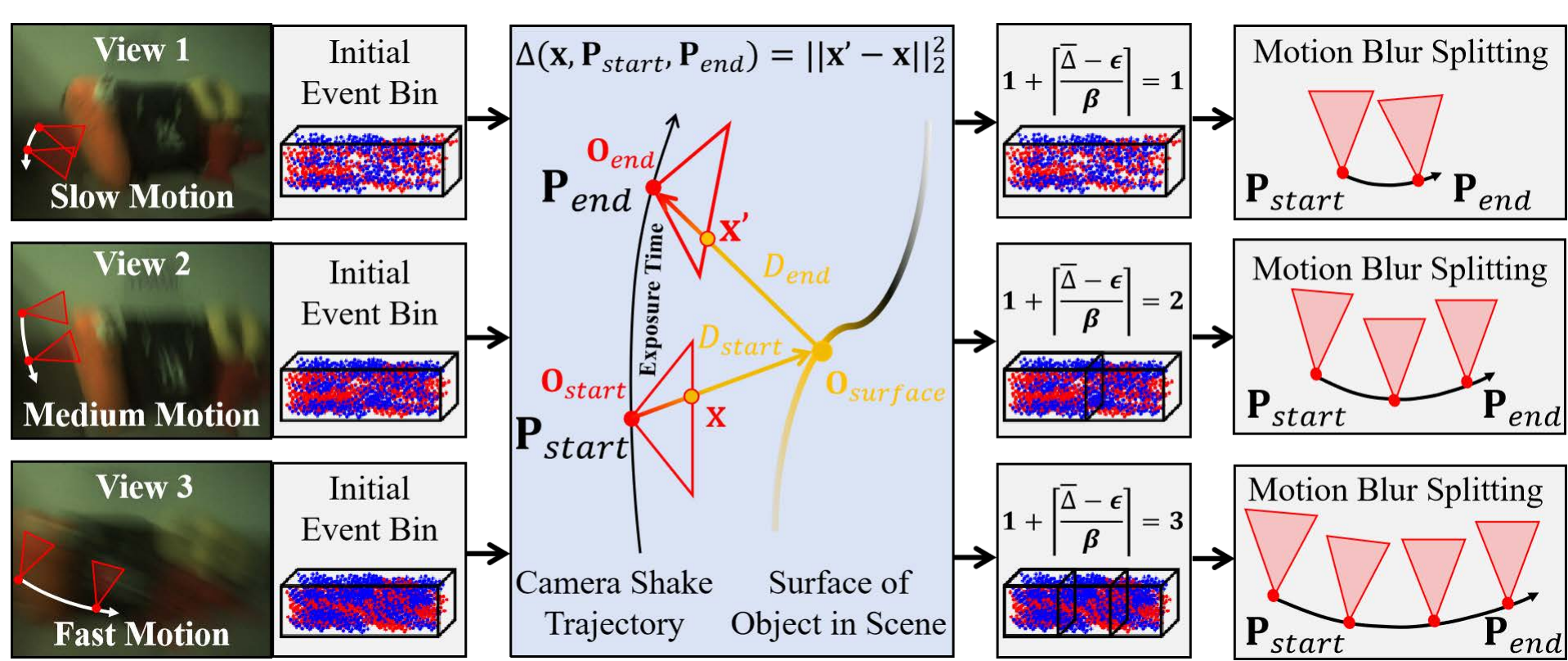}
    \caption{
    Motion-guided event splitting. The left part of the figure is three input images of different views with different degrees of blur and their corresponding events under the same exposure time. Notice that the length of the event bins represents the time, and different densities of events are caused by different motion speeds. The middle of the figure shows the calculation of pixel offset $\Delta$ from pose $\mathbf{P}_{start}$ to $\mathbf{P}_{end}$. The right part of the figure shows the split event bins and camera motion.
    }
    \label{fig:5}
\end{figure}

\subsubsection{Motion-Guided Event Splitting}
\label{sec:4.1.4}

As shown in the left part of Fig.~\ref{fig:5}, input images from different views often exhibit varying degrees of blur.
In E\textsuperscript{2}NeRF, a fixed number of event bins is used across all views.
However, to better accommodate the diverse blur levels in the input images and achieve a balance between efficiency and effectiveness, it is necessary to adapt the number of event bins for each view individually.
Therefore, we compute $\{b_v\}_{v=1}^{V}$ for each view according to the motion blur degree of the corresponding input blurry images.

We first estimate the average pixel offset value for each view.
Specifically, for each pixel $\mathbf{x}=(x,y)$ at pose $\textbf{P}_{start}$ at time $t_{start}$, we can emit a ray $\textbf{r}_{start}$ from camera optical center $\mathbf{o}_{start}$ through the pixel with direction $\mathbf{d}_{start}$ and the corresponding scene surface point is:
\begin{equation}
    \mathbf{o}_{surface}=\mathbf{o}_{start}+D(\mathbf{r}_{start},\mathbf{x})\mathbf{d}_{start},
    \label{eq:13}
\end{equation}
where $D(\mathbf{r}_{start},\mathbf{x})$ is calculated by Eq.~\eqref{eq:4}.
Then, the imaging pixel coordinate $\mathbf{x'}(x',y')$ of the $\mathbf{o}_{surface}$ on the camera at pose $\mathbf{P}_{end}$ at time $t_{end}$ with optical center $\mathbf{o}_{end}$ is:
\begin{equation}
    (x',y',1) = \mathbf{K}\mathbf{R}_{end}\frac{(\mathbf{o}_{surface}-\mathbf{o}_{end})^\mathsf{T}}{|(\mathbf{o}_{surface}-\mathbf{o}_{end})|},
    \label{eq:14}
\end{equation}
where, ${\mathbf{R}}_{end}$ is the camera-to-world rotation matrix at time $t_{end}$ and $\mathbf{K}$ is camera intrinsic matrix.
The pixel offset value at $\mathbf{x}$ during the exposure time from $t_{start}$ to $t_{end}$ is:
\begin{equation}
    \Delta{(\mathbf{x},\mathbf{P}_{start},\mathbf{P}_{end})} = ||({x'}-{x},{y'}-{y})||_2^2,
    \label{eq:15}
\end{equation}
and the average pixel offset of each view is:
\begin{equation}
    \{{\bar\Delta_v}\}_{v=1}^{V} =\frac{1}{|\mathbf{X}_{blur}^{v}|}\sum_{\mathbf{x}\in\mathbf{X}_{blur}^{v}}\Delta{(\mathbf{x},\mathbf{P}_{start},\mathbf{P}_{end})}.
    \label{eq:16}
\end{equation}
$|\mathbf{X}_{blur}^{v}|$ is the number of the blur pixels of the $v$-th input image and $\mathbf{P}_{start}=\mathbf{P}_{0}$, $\mathbf{P}_{end}=\mathbf{P}_{b_{v}}$ of each view can be obtained in Sec.~\ref{sec:4.3}.
Then, we determine an appropriate $b_{v}$ for each view according to the average pixel offset value:
\begin{equation}
    \{b_{v}\}_{v=1}^{V} = \beta (1 + \lceil\frac{\bar\Delta_{v} - \epsilon}{\beta}\rceil),
    \label{eq:17}
\end{equation}
where $\beta$ is an initial bin number and $\epsilon$ is a threshold.
We only activate Eq.~\eqref{eq:17} for severe blurry input views with $\bar\Delta_{v} > \epsilon$.
If $\bar\Delta_{v} \leq \epsilon$, we take $b_{v}=\beta$ directly without splitting initial bins.
We set $\beta=4$ and $\epsilon=6$ in our experiments.

The motion-guided event splitting dynamically splits events of each view into a suitable number of event bins and evenly distributes training across different views, further improving our framework's efficiency and performance.

\subsection{Efficient Event-Enhanced NeRF Network}
\label{sec:4.2}

\subsubsection{Blur Rendering Loss}
\label{sec:4.2.1}

The original NeRF computes the mean squared loss between the predicted images and the input $V$ sharp images across $V$ views.
We modify the NeRF network with a blur rendering loss in this section to fit the blurry inputs of our setting.

For the $v$-th view, we discretize Eq.~\eqref{eq:8} with $b_{v}+1$ virtual sharp frames to simulate the physical formation of the input blurry image $I_{blur}^{v}$:
\begin{equation}
    \begin{aligned}
        {\hat{I}_{blur}^{v}} = \phi\sum_{k=0}^{b_v}I_{vir}(t_{k}),\\
    \end{aligned}
    \label{eq:18}
\end{equation}
where $t_{k}$ is defined in Eq.~\eqref{eq:11}.
With $b_{v}+1$ poses $\{\mathbf{P}_{k}\}_{k=0}^{b_{v}} = \{\mathbf{P}_{t_{k}}\}_{k=0}^{b_{v}}$ at time $t_k$, we can get $b_{v}+1$ rays $\{\mathbf{r}_k\}_{k=0}^{b_{v}}$ for pixel $\mathbf{x}$,
and with Eqs.~\eqref{eq:1} and \eqref{eq:3}, we can get $b_{v}+1$ predicted sharp color values $\{\hat{C}_{k}\}_{k=0}^{b_{v}}=\{C(\mathbf{r}_{k},\mathbf{x})\}_{k=0}^{b_{v}}$.
According to Eq.~\eqref{eq:18}, the predicted blur color of pixel $\mathbf{x}$ is:
\begin{equation}
    \begin{aligned}
        \hat{C'}_{blur}(\mathbf{x}) = \phi\sum_{k=0}^{b_{v}}C(\mathbf{r}_k,\mathbf{x}) = \phi\sum_{k=0}^{b_{v}}C(\mathbf{P}_{t_{k}},\mathbf{x}).\\
    \end{aligned}
    \label{eq:19}
\end{equation}
Since the virtual sharp frames corresponding to $t_k$ are not evenly distributed along the exposure time, we need to multiply a time-based weight parameter:
\begin{equation}
    \{W_{k}\}_{k=0}^{b_{v}} = \frac{t_{k+1} + t_{k-1} - 2t_{k}}{2}
    \label{eq:20}
\end{equation}
for each estimated sharp color to replace the normalization factor $\phi$ in Eq.~\eqref{eq:19}.
We take $t_{-1} = t_{0}$ and $t_{b_{v}+1} = t_{b_{v}}$ and the final predicted blur color is defined as:
\begin{equation}
    \begin{aligned}
        \hat{C}_{blur}(\mathbf{x})=\sum_{k=0}^{b_{v}}W_{k}C(\mathbf{r}_k,\mathbf{x}).\\
    \end{aligned}
    \label{eq:21}
\end{equation}
\textcolor{black}{
Finally, we convert the loss function of NeRF (Eq.~\eqref{eq:5}) into a blur rendering loss:}
\begin{equation}
    \mathcal{L}_{blur} = \sum_{v=1}^{V}\sum_{\mathbf{x}\in\mathbf{X}_{blur}^{v}}[\|\hat{C}^{f}_{blur}(\mathbf{x})-C(\mathbf{x})\|_{2}^{2}+\|\hat{C}_{\textcolor{black}{\kappa}}^{c}(\mathbf{x})-C(\mathbf{x})\|_{2}^{2}].
    \label{eq:22}
\end{equation}
\textcolor{black}{
Note that the first term of Eq.~\eqref{eq:22} represents supervising the blur color $\hat{C}^{f}_{blur}(\mathbf{x})$ predicted by the fine network and the second term represents supervising the color $\hat{C}_{\kappa}^{c}(\mathbf{x})$ predicted by the coarse network at a randomly selected pose $\mathbf{P}_\kappa \in \{\mathbf{P}_{t_{k}}\}_{k=0}^{b_{v}}$, which is to impose a strong constraint on the fine network and a weak constraint on the coarse network, separately, improving the stability of training.}

\subsubsection{Event Rendering Loss}
\label{sec:4.2.2}

The blur rendering loss can not ensure \textcolor{black}{the values of $b_{v}+1$ predicted intermediate virtual sharp images are correct, because it is only supervised by the input blurry image for each view, which only guarantees the accuracy of their weighted average.}
Leveraging the high temporal resolution intensity change information in event data, we introduce the event rendering loss, which further supervises the values of these virtual sharp images and makes up for the limitations of the blur rendering loss in modeling the blurring process.

\textcolor{black}{
For the $v$-th view, given a blurred pixel $\mathbf{x}(x,y) \in \mathbf{X}_{blur}$, we first select two estimated color values of the adjacent virtual sharp images ${\hat{C}}_{k}, {\hat{C}}_{k-1}$ at time $t_k$, $t_{k-1}$.
Then we convert them into gray-scale values as $L_{k}$, $L_{k-1}$, take the absolute difference between the two values in the log domain, and divide it by the threshold $\Theta$ to estimate the absolute event number at pixel $\mathbf{x}$}:
\begin{equation}
    {\rm sum} (\hat{B}_{k}(\mathbf{x}))=\left\{
    \begin{aligned}
    &\lceil{\frac{\log({L_{k-1})} - \log({L_{k}})}{\Theta_{neg}}\rceil}, {L_{k}}<{L_{k-1}}\\
    &\lfloor{\frac{\log({L_{k})} - \log({L_{k-1}})}{\Theta_{pos}}\rfloor},{L_{k}}\geq{L_{k-1}}
    \end{aligned}.
    \right.
    \label{eq:23}
\end{equation}
\textcolor{black}{
Note that we use ${\rm sum}(\hat{B}_{k}(\mathbf{x}))$ to represent setting the number of negative events to its additive inverse and adding it to the number of positive events to get the absolute event number of event bin $\hat{B}_{k}(\mathbf{x})$, which can be calculated by the right side of the Eq.~\eqref{eq:23} directly.
Finally, we use the mean squared error between these estimated absolute event numbers of event bins $\{\hat{B}_{k}(\mathbf{x})\}_{k=1}^{b_v}$ and those of input event bins $\{B_{k}(\mathbf{x})\}_{k=1}^{b_v}$ as our event rendering loss:}
\begin{equation}
    \begin{aligned}
    \mathcal{L}_{event} = \sum_{v=1}^{V}\sum_{\mathbf{x}\in\mathbf{X}_{blur}^{v}} \sum_{k=1}^{b_{v}}\|{\rm sum}(\hat{B}_{k}(\mathbf{x}))-{\rm sum}(B_{k}(\mathbf{x}))\|_{2}^{2}.
    \end{aligned}
    \label{eq:24}
\end{equation}
\textcolor{black}{
Note that we replace the event estimation between two random virtual sharp frames in E\textsuperscript{2}NeRF with two adjacent virtual sharp frames.
This can avoid the large changes in the number of events between the two selected frames that conflict with the temporal blur uniform binning in Sec.~\ref{sec:4.1.2}.}
As a result, the variation of ${L}_{event}$ is significantly reduced, and the network is more stable during the training.

\subsubsection{Final Loss}
\label{sec:4.2.3}

Unlike E\textsuperscript{2}NeRF computing $\mathcal{L}_{event}$ and $\mathcal{L}_{blur}$ for all pixels, E\textsuperscript{3}NeRF only conducts these two computationally expensive losses for the spatial blurred pixels $\mathbf{X}_{blur}$ located by Sec.~\ref{sec:4.1.3}.
As for the sharp pixels $\mathbf{X}_{sharp}$ we randomly select one pose $\mathbf{P}_\kappa \in \{\mathbf{P}_{t_{k}}\}_{k=0}^{b_{v}}$ to render the sharp color $\hat{C}_{sharp}(\mathbf{x})$ and calculate $\mathcal{L}_{sharp}$ with Eq.~\eqref{eq:5} as in the original NeRF.
This concentrates training resources on regions with blur and regions containing texture details in the 3D scene.
Finally, our final loss function is defined as:
\begin{equation}
    \mathcal{L}=\mathcal{L}_{blur} + \mathcal{L}_{sharp} + \lambda\mathcal{L}_{event},
    \label{eq:25}
\end{equation}
where $\lambda$ is the weight parameter of $\mathcal{L}_{event}$.

\subsection{Event-Guided Pose Estimation}
\label{sec:4.3}

Generally, NeRF uses the camera poses estimated by COLMAP\cite{colmap} for real-world data.
However, when the input image becomes severely blurry, the pose estimation results of COLMAP will become inaccurate or even invalid, limiting NeRF's application in low-light and high-speed real scenes of our settings.
Besides, we need $b_{v}+1$ poses $\{\mathbf{P}_{k}\}_{k=0}^{b_{v}}$ at $\{t_{k}\}_{k=0}^{b_{v}}$ for the $v$-th view to render corresponding virtual sharp images $\{I_{k}\}_{k=0}^{b_{v}}$ for the blur rendering loss and event rendering loss.
Therefore, we design an event-guided pose estimation framework for real-world data by utilizing the high temporal resolution information in events.

Specifically, for the $v$-th view with an input blurry image $I_{blur}^{v}$ and the corresponding event bins $\{B_{k}\}_{k=1}^{b_{v}}$, according to the event-based double integral model (EDI) \cite{pan2019bringing}, the virtual sharp images at time $\{t_{k}\}_{k=0}^{b_{v}}$ can be expressed as:
\begin{equation}
    \{I_{k}\}_{k=1}^{b_{v}} = I_{0}e^{\Theta\sum_{j=1}^{k}{{\rm sum}(B_{j})}},(k=1,2,...,b_{v}),
    \label{eq:26}
\end{equation}
where $I_{0}$ is the virtual sharp image at $t_0$.
Then according to the image formation model in Eq.~\eqref{eq:8} and time-based weight in Eq.~\eqref{eq:20}, the blurry image can be expressed as:
\begin{equation}
\begin{aligned}
    I_{blur}^{v} = \sum_{k=0}^{b_{v}}{W_k}{I_k} = I_0(W_0 + W_1e^{\Theta\sum_{j=1}^{1}{{\rm sum}(B_{j})}} \\ 
     + \dots + W_{b_{v}}e^{\Theta\sum_{j=1}^{b_{v}}{{\rm sum}(B_{j})}}),
    \end{aligned}
\label{eq:27}
\end{equation}
and $I_{0}$ can be transformed into:
\begin{equation}
    I_{0} = \frac{I_{blur}^{v}}{W_0+ W_1e^{\Theta\sum_{j=1}^{1}{{\rm sum}(B_{j})}}+\dots+W_{b_{v}}e^{\Theta\sum_{j=1}^{b_{v}}{{\rm sum}(B_{j})}}}.
    \label{eq:28}
\end{equation}
Substituting Eq.~\eqref{eq:28} into Eq.~\eqref{eq:26} we can get $\{I_{k}\}_{k=1}^{b_{v}}$ calculated by the input blurry image and events.
We feed $\{I_{k}\}_{k=0}^{b_{v}}$ into COLMAP to get $b_{v}+1$ poses for the $v$-th view:
\begin{equation}
    \{\mathbf{P}_{k}\}_{k=0}^{b_{v}} = \text{COLMAP}(\{I_{k}\}_{k=0}^{b_{v}}).
    \label{eq:29}
\end{equation}
Finally, we do the same operation for all the input views synchronously to generate all the poses for training.

The pose estimation framework uses event data to enhance the robustness of pose estimation against real-world data characterized by severe and non-uniform motion blur, generalizing our method to more practical applications.

\section{Experiments}
\label{sec:5}

\subsection{Datasets}
\label{sec:5.1}

\subsubsection{Synthetic Data}
\label{sec:5.1.1}
We use Blender \cite{Blender} to generate two sets of synthetic data with slightly and severely blurry input, respectively.
The datasets consist of seven scenes in NeRF \cite{NeRF} (Chair, Drums, Ficus, Hotdog, Lego, Materials, and Mic).
Each scene consists of 100 views of blurry images and the corresponding events for training, and 100 novel views of sharp images for testing.
We also synthesize 100 sharp images corresponding to the input blurry images to evaluate the image deblurring effect.
We use the ``Camera Shakify Plugin'' of Blender to simulate handheld camera shake and render $n$ sharp images during the shaking. 
The sharp images are used to synthesize the blurry images and to generate the events through v2e \cite{esim2} with ``noisy'' option and thresholds $\Theta_{pos}=\Theta_{neg}=0.25$.
We use $n=17$ and $n=33$ for the slightly and severely blurry datasets, respectively.
In addition, we randomly adjust the camera shake speed for the severely blurry data to introduce the non-uniform motion as described in Fig.~\ref{fig:3}.
The detailed generation of blurry images is shown in the supplementary materials.

\begin{table*}[t]
\caption{Comparision bewteen Real-World-Blur dataset and Real-world-Challenge dataset. The upper part of the table shows the blur degrees of the input images.
The lower part of the table shows the number of poses successfully generated by COLMAP \cite{colmap} and our pose estimation framework.}
\label{tab:2}
    \setlength{\tabcolsep}{1.5mm}
    \centering
    \begin{tabular}{c|c||ccccc|c||ccccc|c}
        \toprule
        & & \multicolumn{6}{c||}{Real-World-Blur Dataset (30 Input Training Views)} & \multicolumn{6}{c}{Real-World-Challenge Dataset (16 Input Training Views)} \\
        & & Camera & Lego & Letter  & Plant & Toys & Mean & Corridor & Lab & Lobby & Shelf & Table & Mean \\
        \midrule
        \multirow{3}*{\shortstack{Blur Range\\ (pixel)}} & ${\bar{\Delta}}_{min}$  & 0.74 & 0.25 & 0.40 & 0.30 & 0.33 & 0.40 & 1.99  & 3.37  & 2.46  & 3.39  & 3.20  & 2.88 \\
        & ${\bar{\Delta}}_{max}$  & 7.30 & 4.24 & 7.46 & 5.47 & 8.22 & 6.54 & 11.09 & 14.46 & 13.51 & 13.91 & 11.50 & 12.89 \\
        & ${\bar{\Delta}}_{ave}$  & 2.37 & 1.68 & 2.32 & 1.83 & 3.24 & 2.29 & 5.39 & 8.19   & 5.53  & 7.80  & 6.69  & 6.72 \\
        \midrule
        \multirow{2}*{\shortstack{Number of \\ Estimated Poses}} 
        & COLMAP         & 14 & 25 & 25 & 27 & 24 & 23 & 0 & 0 & 0 & 0 & 0 & 0 \\ 
        & Ours  & 30 & 30  & 30 & 30 & 30 & 30 & 16 & 16 & 16 & 16 & 16 & 16 \\ 
        \bottomrule
        \end{tabular}
        \begin{tablenotes}
        \item ${\bar{\Delta}}$ is calculated by Eq.~\eqref{eq:16}. ${\bar{\Delta}}_{min}$: Minimum ${\bar{\Delta}}$ of all views. ${\bar{\Delta}}_{max}$: Maximum ${\bar{\Delta}}$ of all views. ${\bar{\Delta}}_{ave}$: Average ${\bar{\Delta}}$ of all views.
    \end{tablenotes}
\end{table*}

\subsubsection{Real-World Data}
\label{sec:5.1.2}
We construct two sets of real-world datasets with different ranges of blur by DAVIS 346 color event camera\cite{davis346}.
The camera is capable of capturing spatial-temporal aligned event data and RGB images with a resolution of 346$\times$260.
Each dataset consists of five low-light scenes with illumination ranging from 5 to 100 lux.
TABLE~\ref{tab:2} shows a comparison between these two datasets on blur range and the pose estimation results of COLMAP and our framework.

\begin{figure}[t]
    \centering
    \includegraphics[width=1\linewidth]{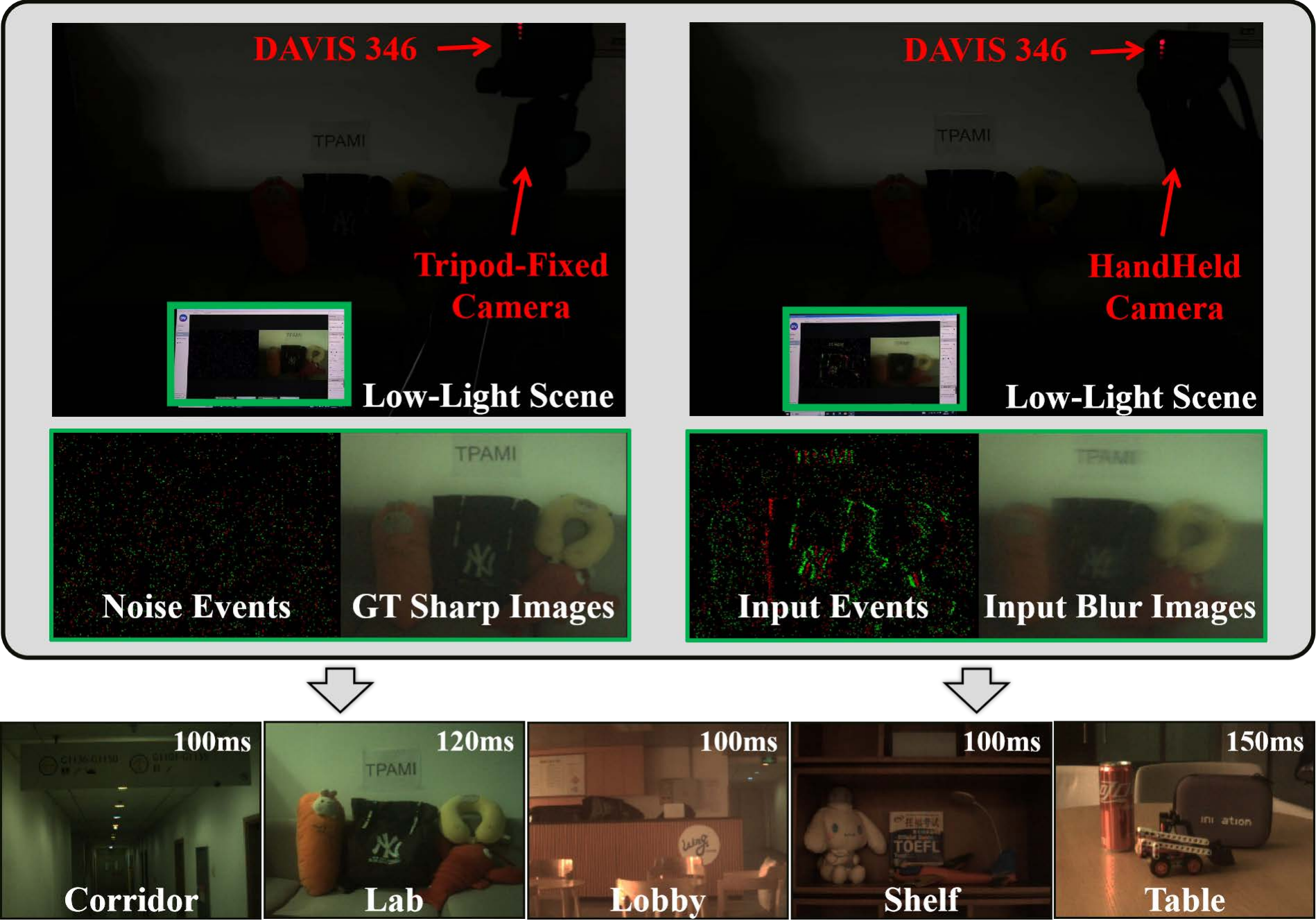}
    \caption{
    Collection of the Real-World-Challenge dataset.
    The training data and ground-truth test data are captured with a handheld and tripod-fixed DAVIS 346 camera, respectively.
    The five scenes with varying lighting conditions and scales are illustrated at the bottom of the figure. The exposure times are marked on the upper right of the images.}
    \label{fig:6}
\end{figure}

\begin{figure}[t]
    \centering
    \color{black}
    \includegraphics[width=1\linewidth]{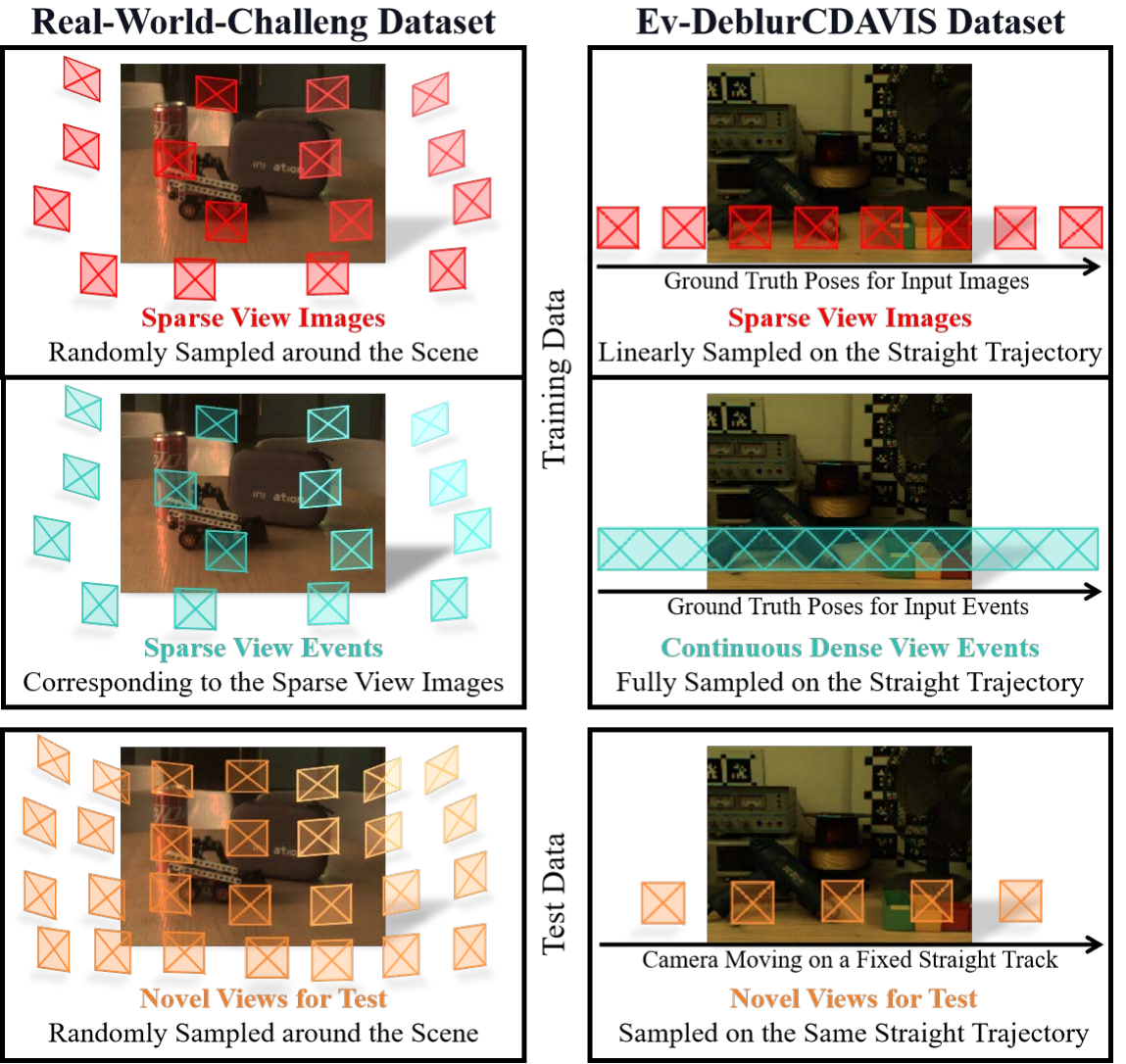}
    \caption{Comparison between our Real-World-Challenge dataset and Ev-DeblurCDAVIS dataset. Red, cyan, and orange represent the input image and event views, and the novel views for test, respectively. Our dataset is more challenging in four aspects: 1. The input sparse view images in our dataset are randomly sampled around the scene, rather than linearly sampled along a straight trajectory; 2. The input sparse view events in our dataset correspond only to the input image, which are significantly fewer than the continuous dense view events fully sampled along the straight trajectory. 3. The Ev-DeblurCDAVIS dataset provides ground-truth poses for both input images and events, whereas ours does not. 4. The test novel views of the Ev-DeblurCDAVIS dataset are sampled on the same straight trajectory and overlap with the input event views, while ours are randomly sampled and completely novel.}
    \label{fig:7}
\end{figure}

\myPara{Real-World-Blur Dataset:}
This dataset is captured by a handheld DAVIS 346 and consists of five scenes: Camera, Lego, Letter, Plant, and Toys.
The exposure time is 100 ms for the RGB sensor to capture images with sufficient brightness in low-light scenes, which also results in image blur.
Each scene has 30 different views of blurry images and the corresponding event data for training.

\myPara{Real-World-Challenge Dataset:}
This dataset comprises five challenging scenes: Corridor, Lab, Lobby, Shelf, and Table, which cover various lighting conditions and scene scales.
The exposure times of each scene are shown in Fig.~\ref{fig:6}.\
We capture the training data with a handheld DAVIS 346 and the ground-truth test data with a tripod-fixed DAVIS 346.
Each scene consists of 16 randomly sampled views of blurry images and corresponding events for training, and 28 novel view sharp images for testing, as shown in the left part of Fig.~\ref{fig:7}.
We also amplify the camera motion to obtain more challenging training data for a more comprehensive evaluation of the model's robustness to severe blur.
As shown in the TABLE~\ref{tab:2}, the maximum, minimum, and average blur ranges of the Real-World-Challenge dataset are all larger than those of the Real-World-Blur dataset, which leads to a complete failure of COLMAP's pose estimation on the Real-World-Challenge dataset. 
In contrast, our framework successfully estimates all poses for all scenes of both datasets.

\subsubsection{Public Datasets}
\label{sec:5.1.3}
As described in Sec.~\ref{sec:2.3} and TABLE~\ref{tab:1}, previous event-based NeRF and ERGB-based deblurring NeRF methods all need extra events and ground-truth poses as input, which are not provided in our proposed synthetic and real-world data.
Therefore, we introduce two public datasets containing such data for further comparison between our E\textsuperscript{3}NeRF and existing event-related NeRF works.

\myPara{TUM-VIE Dataset \cite{tumvie}:}
This dataset is used in E-NeRF \cite{enerf2}, and \emph{e}-NeRF \cite{regnerf}, containing continuous dense view events and high-frequency acquired ground-truth camera poses.
It also contains images as input for our E\textsuperscript{3}NeRF and ground-truth images for evaluation.
We utilize two scenes, ``mocap-1d-trans'' and ``mocap-desk2'', of this dataset in our experiment, as in E-NeRF and \emph{e}-NeRF.

\myPara{Ev-DeblurCDAVIS Dataset \cite{evdeblurnerf}:}
\textcolor{black}{
This dataset is proposed in Ev-DeblurNeRF \cite{evdeblurnerf} and contains five real scenes, captured by a DAVIS 346 event camera moving along a straight fixed track in front of the scenes as shown in the right part of Fig.~\ref{fig:7}.
The sparse view images for training are linearly sampled on the straight camera motion trajectory, and the continuous dense view events for training are fully sampled on the straight trajectory.
The ground-truth poses are also provided at high frequency.
The test novel views are sampled on the same straight trajectory and overlap with the input event views.
Fig.~\ref{fig:7} shows a comparison between our Real-World-Challenge dataset and the Ev-DeblurCDAVIS dataset.
Our dataset is more challenging in terms of input view and test view distribution, and has less input data (no ground-truth poses and extra dense view events).
}

\subsection{Compared Methods}
\label{sec:5.2}

\myPara{Image-Based Deblurring NeRF Methods:}
We compare our E\textsuperscript{3}NeRF with three image-based deblurring NeRF methods: Deblur-NeRF \cite{Deblur-NeRF}, BAD-NeRF \cite{bad-nerf}, and DP-NeRF \cite{dpnerf}.
We also select the state-of-the-art single-image deblurring method MPR\cite{sdeblur7} and event-enhanced image deblurring methods D2Net\cite{shang2021bringing} and EDI\cite{pan2019bringing} for comparison.
Additionally, we train NeRF with the images deblurred by the aforementioned methods and name them MPR-NeRF, D2Net-NeRF, and EDI-NeRF to compare them with our method.

\myPara{Event-Based NeRF Methods:}
Some of the mentioned event-based NeRF methods in Sec.~\ref{sec:2.3} are difficult to apply in our experiments.
For example, there is no open-source code available for the Ev-NeRF \cite{enerf1}.
EventNeRF \cite{eventnerf} can only be used for its real-world dataset that is collected by objects uniformly rotating on a turntable with a white background.
We choose E-NeRF with only events as input and \emph{e}-NeRF as the event-based NeRF methods for comparison and conduct the experiment on the TUM-VIE\cite{tumvie} dataset with ground-truth poses and continuous dense view events as input.

\myPara{ERGB-Based NeRF Methods:}
As described in Sec.~\ref{sec:2.3} and TABLE~\ref{tab:1}, DE-NeRF \cite{denerf} aims to reconstruct a dynamic deformable NeRF from ERGB data, and the code has not been open-sourced.
Be-NeRF \cite{benerf} trains separate NeRFs for each view, which differs from our settings and struggles in novel view synthesis on unseen views.
Therefore, we select the concurrent ERGB-based NeRF work, E-NeRF \cite{enerf2}, and the latest ERGB-based deblurring work, Ev-DeblurNeRF \cite{evdeblurnerf}, for comparison.
Since E-NeRF and Ev-DeblurNeRF are not designed for Blender data with 360° input views and require continuous dense view events that are not only corresponding to the image views but also between the image views as training data, we first train Ev-DeblurNeRF on our Real-World-Challenge dataset following our settings without ground-truth pose and extra events for a basic comparison.
Then, we compare these two methods with our E\textsuperscript{3}NeRF in detail on the EvDeblurCDAVIS dataset \cite{evdeblurnerf} for a more comprehensive comparison.

\subsection{Implementation Details}
\label{sec:5.3}

\myPara{E\textsuperscript{3}NeRF:}
Our code is based on NeRF.
We set $\lambda=0.005$, $N_{coarse}=64$, $N_{fine}=128$ for the coarse and fine network respectively.
For synthetic data and Real-World-Blur data, we use a batch size of 1024 and train each scene for 200,000 iterations, which is the same as in E\textsuperscript{2}NeRF \cite{e2nerf}.
For the Real-World-Challenge dataset, Ev-DeblurCDAVIS dataset, and TUM-VIE dataset, we set the batch size to 512 and train for 50,000 iterations, as the number of input views is reduced.
We set the positive and negative thresholds to $0.25$ for synthetic data, which matches the settings of the event simulation process in Sec.~\ref{sec:5.1.1}.
For the real-world data, we set the positive and negative thresholds to 0.3, a middle value of the threshold distribution range from 0.1 to 0.5 of event sensors \cite{eventsurvey}.
In the first 10,000 iterations, we set $b=4$ for all views to pre-train the network and get a basic 3D structure of the scene.
Then, we calculate $b_{v}$ for each view with the motion-guided event splitting model (Sec.~\ref{sec:4.1.4}) and use it in the subsequent iterations.
All experiments are implemented on a single NVIDIA RTX 3090 GPU.

\myPara{\textcolor{black}{Compared Methods:}}
\textcolor{black}{We use the same NeRF-related settings for all compared NeRF-based methods to ensure a fair comparison.
For event-based and ERGB-based methods, we also used the same event-related parameters.
The codes and parameters for the experiments on public datasets are all derived from their public official implementation files.
}

\begin{table*}[t]
    \caption{Quantitative results of blur view and novel view on the synthetic data. The results are the averages of the seven synthetic scenes.}
    \label{tab:3}
    \centering
    \footnotesize
    \setlength{\tabcolsep}{0.5mm}
    \begin{tabular}{c|c|c||ccc|cccccccccc}
        \toprule
            \multicolumn{3}{c||}{} & \multicolumn{3}{c|}{25 Forward Facing Views} & \multicolumn{9}{c}{100 Full 360° Views}\\
        \midrule
            & Datasets & Metrics & Deblur-NeRF$^{25}$ & BAD-NeRF$^{25}$ & E\textsuperscript{3}NeRF$^{25}$ & NeRF & D2Net& D2Net-NeRF & EDI & EDI-NeRF & MPR & MPR-NeRF & E\textsuperscript{2}NeRF & E\textsuperscript{3}NeRF\\
        \midrule
            \multirow{6}*{\rotatebox[origin=c]{90}{Blur View}} & \multirow{3}*{\shortstack{Slight \\ Shake}}
            & PSNR\textuparrow     & 21.56 & 12.68 & \textbf{32.56} & 22.30 & 27.04 & 26.94 & 28.74 & 29.43 & 27.41 & 27.23 & 30.65 & \textbf{31.41} \\
            & & SSIM\textuparrow     & .8755 & .7814 & \textbf{.9763} & .8991 & .9449 & .9409 & .9566 & .9635 & .9497 & .9467 & .9690 & \textbf{.9713} \\
            & & LPIPS\textdownarrow  & .2437 & .4137 & \textbf{.0349} & .1564 & .0983 & .1098 & .0765 & .0573 & .0928 & .0975 & .0497 & \textbf{.0412} \\
        \cmidrule(lr){2-15}
            & \multirow{3}*{\shortstack{Severe \\ Shake}}
            & PSNR\textuparrow     & 18.85 & 12.44 & \textbf{29.62} & 21.05 & 21.22 & 21.33 & 26.36 & 26.83 & 21.73 & 22.38 & 26.82 & \textbf{29.12} \\ 
            & & SSIM\textuparrow     & .8483 & .7945 & \textbf{.9649} & .8885 & .8889 & .8898 & .9453 & .9594 & .9001 & .9045 & 9515 & \textbf{.9627} \\
            & & LPIPS\textdownarrow  & .2899 & .4274 & \textbf{.0555} & .2085 & .1775 & .1928 & .1340 & .0687 & .1651 & .1635 & 0918 & \textbf{.0548} \\
        \midrule
            \multirow{6}*{\rotatebox[origin=c]{90}{Novel View}} & \multirow{3}*{\shortstack{Slight \\ Shake}}
            & PSNR\textuparrow      & 19.66 & 12.66 & \textbf{31.95} & 21.67 & - & 26.70 & - & 29.13 & - & 26.96 & 30.24 & \textbf{31.15} \\
            & & SSIM\textuparrow      & .8516 & .7815 & \textbf{.9750} & .8932 & - & .9406 & - & .9631 & - & .9457 & .9689 & \textbf{.9712} \\
            & & LPIPS\textdownarrow   & .2656 & .4161 & \textbf{.0371} & .1610 & - & .1122 & - & .0597 & - & .0986 & .0507 & \textbf{.0426} \\
        \cmidrule(lr){2-15}
            & \multirow{3}*{\shortstack{Severe \\ Shake}}
            & PSNR\textuparrow      & 18.31 & 12.33 & \textbf{29.28} & 21.00 & - & 21.28 & - & 26.87 & - & 21.81 & 26.69 & \textbf{28.97} \\
            & & SSIM\textuparrow      & .8410 & .7892 & \textbf{.9629} & .8877 & - & .8890 & - & .9595 & - & .9009 & .9508 & \textbf{.9619} \\
            & & LPIPS\textdownarrow   & .3021 & .4295 & \textbf{.0570} & .2102 & - & .1942 & - & .0677 & - & .1714 & .0929 & \textbf{.0556} \\
        \bottomrule
    \end{tabular}
\end{table*}

\subsection{Results of Synthetic Data}
\label{sec:5.4}

We divide the experimental results of synthetic data into two groups: blur view and novel view.
Blur view corresponds to the input blurry images, while novel view has no input images for reference.
We only show the results of blur views for the image deblurring methods because they are not designed for novel view synthesis.
Since Deblur-NeRF and BAD-NeRF can not learn the implicit representation for scenes with 360° views as input, we only use 25 forward-facing views as input for them to ensure a basic reconstruction result.
Accordingly, we also use the same input for E\textsuperscript{3}NeRF for a fair comparison and name them as Deblur-NeRF$^{25}$, BAD-NeRF$^{25}$, and E\textsuperscript{3}NeRF$^{25}$.
The poses are given by Blender as in NeRF\cite{NeRF} for all the methods.

\subsubsection{Quantitative Results}
\label{sec:5.4.1}
We evaluate the results quantitatively with PSNR, SSIM, and LPIPS\cite{lpips}.
Our method achieves the best results on the three metrics and has a significant improvement over all the compared methods on both blur and novel view experiments, as shown in TABLE~\ref{tab:3}.
Though we use forward-facing views as input for Deblur-NeRF and BAD-NeRF, their performance is still inferior because of the failure of the joint learning of the blur kernel and motion blur trajectory, separately.
E\textsuperscript{3}NeRF$^{25}$ with only 25 views of 180° of the scene as input has a slight performance improvement over E\textsuperscript{3}NeRF.
This is because the forward-facing views are more accessible than the 360° views in 3D implicit learning.
The results also show that EDI-NeRF is better than EDI. This is because the performance of EDI is affected by the noise in the deblurred images, and the NeRF training process can weaken this adverse impact of the noise, as proved in RawNeRF \cite{nerfind}.
D2Net, D2Net-NeRF, MPR, and MPR-NeRF show visible improvement on slight shake data compared to the original NeRF, but their performance drops obviously on the severe shake data.
Notice that on the slight shake datasets, the results of E\textsuperscript{2}NeRF and E\textsuperscript{3}NeRF are very close.
But the performance gap between E\textsuperscript{2}NeRF and E\textsuperscript{3}NeRF is widened on the severe shake data on both blur and novel views, proving that the proposed event-guided spatial-temporal blur model significantly strengthens the robustness of E\textsuperscript{3}NeRF.

\begin{figure*}[t]
\centering
    \includegraphics[width=1\linewidth]{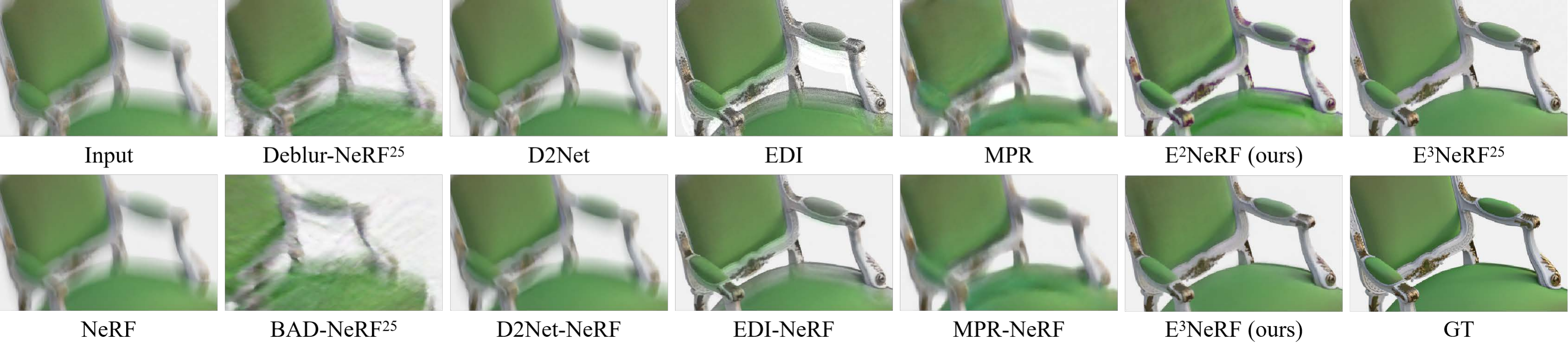}
    \caption{Qualitative results of synthetic data ``Chair'' scene on blur views. Our method has the sharpest result without color deviation and noise.}
    \label{fig:8}
\end{figure*}

\begin{figure*}[ht]
    \centering
    \includegraphics[width=1\linewidth]{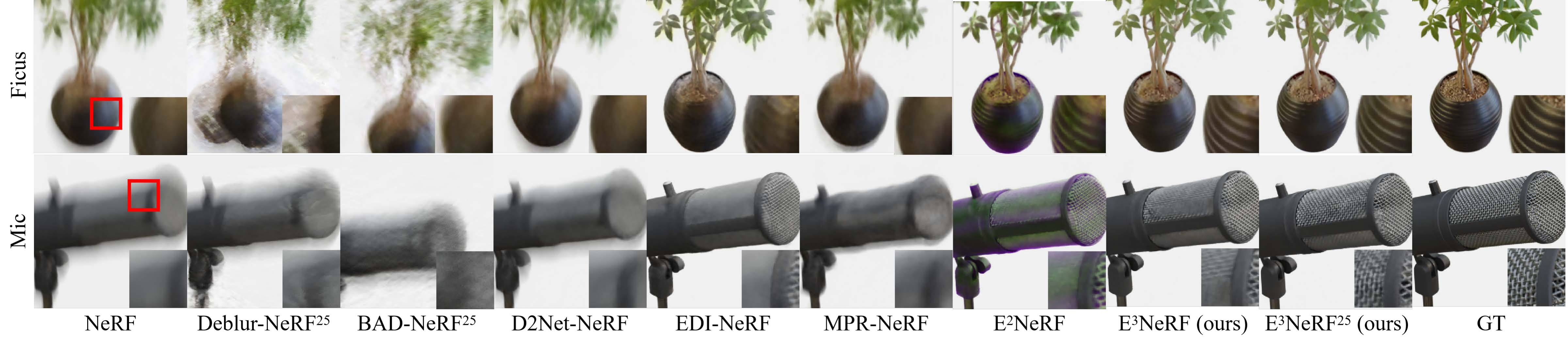}
    \caption{Qualitative results on ``Ficus'' and ``Mic'' scene of synthetic data on novel views. Our method reconstructs the wrinkles on the ficus pot and the mesh structure on the mic from severely blurry input.}
    \label{fig:9}
\end{figure*}

\subsubsection{Qualitative Results}
\label{sec:5.4.2}

\myPara{Blur View:}
In Fig.~\ref{fig:8} we show the rendering results of the ``Chair'' scene on a blur view of the severe shake dataset.
The qualitative results are highly consistent with the quantitative results.
Deblur-NeRF and BAD-NeRF have the worst results due to the misestimation of the blur kernel and camera motion trajectories.
The results of D2Net, D2Net-NeRF, MPR, and MPR-NeRF still have significant blur.
Although EDI and EDI-NeRF produce very sharp results, there is a significant color deviation at the edges of objects.
E\textsuperscript{3}NeRF eliminates the ghosting phenomenon at the edge of E\textsuperscript{2}NeRF and achieves the result closest to the ground truth.

\myPara{Novel View:}
In Fig.~\ref{fig:9}, we show the rendering results of ``Ficus'' and ``Mic'' scenes on a novel view of the severe shake dataset.
With the help of event data, EDI-NeRF, E\textsuperscript{2}NeRF, and E\textsuperscript{3}NeRF successfully recover the wrinkles on the ficus pot and the mesh structure on the mic, while other methods only show the blurred rendering results without these texture details.
When it comes to the more challenging mesh structure on the side of the mic, E\textsuperscript{3}NeRF is significantly better than EDI-NeRF and E\textsuperscript{2}NeRF. 

\begin{table*}[t]
    \caption{Quantitative analysis on Real-World-Blur dataset. The results are the averages of five scenes on both blur view and novel view.}
    \label{tab:4}
    \centering
    \begin{tabular}{c||cccccc|ccc}
        \toprule
        Blur View \& Novel View & NeRF & D2Net-NeRF & MPR-NeRF & EDI-NeRF & Deblur-NeRF & BAD-NeRF & E\textsuperscript{2}NeRF & E\textsuperscript{3}NeRF \\
        \midrule
        RankIQA\textdownarrow   & 5.464 & 4.693 & 4.563 & 3.936 & 4.165 & 4.379 & 3.609 & \textbf{3.243} \\
        MetaIQA\textuparrow     & .1887 & .1893 & .1880 & .2181 & .2067 & .1908 & .2160 & \textbf{.2438} \\      
        \bottomrule
    \end{tabular}
\end{table*}

\begin{figure*}[t]
    \centering
    \includegraphics[width=1\linewidth]{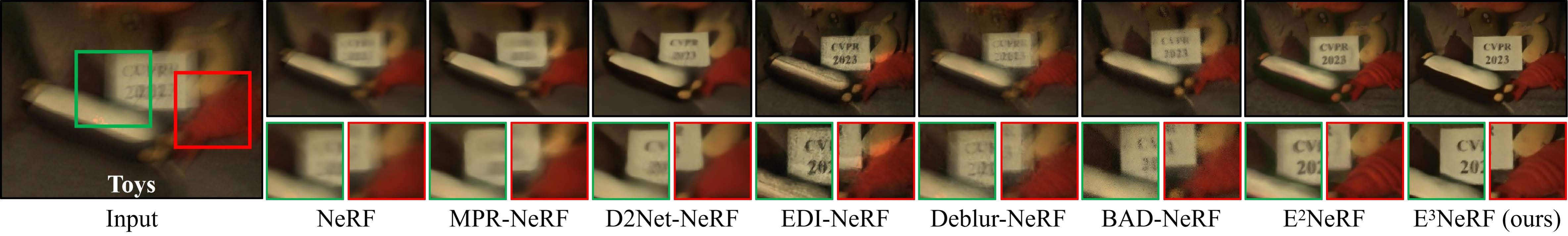}
    \caption{Qualitative results on ``Toys'' scene of Real-World-Blur dataset.}
    \label{fig:10}
\end{figure*}

\begin{table*}[t]
    \caption{Quantitative analysis on the five scenes of the Real-World-Challenge dataset.}
    \label{tab:5}
    \centering
    \footnotesize
    \setlength{\tabcolsep}{0.1mm}
    \begin{tabular}{l|ccc|ccc|ccc|ccc|ccc|ccc}
        \toprule
        & \multicolumn{3}{c}{Corridor} & \multicolumn{3}{c}{Lab}  & \multicolumn{3}{c}{Lobby} & \multicolumn{3}{c}{Shelf} & \multicolumn{3}{c}{Table} & \multicolumn{3}{c}{Average}\\
        
        & PSNR$\uparrow$ & SSIM$\uparrow$ & LPIPS$\downarrow$ & PSNR$\uparrow$ & SSIM$\uparrow$ & LPIPS$\downarrow$ & PSNR$\uparrow$ & SSIM$\uparrow$ & LPIPS$\downarrow$ & PSNR$\uparrow$ & SSIM$\uparrow$ & LPIPS$\downarrow$ & PSNR$\uparrow$ & SSIM$\uparrow$ & LPIPS$\downarrow$ & PSNR$\uparrow$ & SSIM$\uparrow$ & LPIPS$\downarrow$ \\
        \midrule
        NeRF         & 27.88 & .9341 & .4446 & 27.27 & .9045 & .4229 & 24.58 & .8705 & .4978 & 26.67 & .8463 & .4619 & 24.85 & .8766 & .4302 & 26.25 & .8864 & .4515 \\  
        MPR-NeRF     & 28.26 & .9375 & .4058 & 28.07 & .9144 & .3655 & 26.58 & .8935 & .4088 & 25.69 & .8116 & .4617 & 25.33 & .8882 & .3652 & 26.79 & .8890 & .4014 \\
        D2Net-NeRF   & 28.17 & .9376 & .4127 & 27.83 & .9100 & .3771 & 26.78 & .8956 & .4032 & 26.67 & .8501 & .4261 & 25.03 & .8811 & .3959 & 26.90 & .8949 & .4030 \\    
        EDI-NeRF     & \textbf{31.81} & .9570 & .2865 & 27.64 & .9267 & .2929 & 28.05 & .9197 & .3027 & 29.66 & .9125 & .2167 & 27.28 & .9218 & .2499 & 28.89 & .9275 & .2697 \\
        \midrule
        
        Deblur-NeRF  & 27.76 & .9337 & .3996 & 28.44 & .9130 & .3575 & 25.15 & .8756 & .4466 & 26.20 & .8411 & .4625 & 25.17 & .8795 & .3950 & 26.54 & .8886 & .4122 \\    
        BAD-NeRF   & 26.24 & .8950 & .4761 & 28.86 & .9125 & .2386 & 23.06 & .8175 & .6265 & 29.61 & .8820 & .2487 & 20.39 & .7804 & .6102 & 25.63 & .8575 & .4400 \\
        \color{black}{DP-NeRF} & \color{black}{29.64} & \color{black}{.9508} & \color{black}{.3361} & \color{black}{31.40} & \color{black}{.9452} & \color{black}{.2350} & \color{black}{27.52} & \color{black}{.9168} & \color{black}{.3093} & \color{black}{29.92} & \color{black}{.9036} & \color{black}{.2937} & \color{black}{25.79} & \color{black}{.8964} & \color{black}{.3336} & \color{black}{28.85} & \color{black}{.9226} & \color{black}{.3015} \\
        \midrule
        
        \color{black}{Ev-DeblurNeRF*}  & \color{black}{31.27} & \color{black}{.9546} & \color{black}{.2960} & \color{black}{31.64} & \color{black}{.9570} & \color{black}{.1790} & \color{black}{22.48} & \color{black}{.8611} & \color{black}{.4917} & \color{black}{31.53} & \color{black}{.9285} & \color{black}{.2141} & \color{black}{22.23} & \color{black}{.8647} & \color{black}{.3975} & \color{black}{27.83} & \color{black}{.9132} & \color{black}{.3157} \\
        E\textsuperscript{2}NeRF   & 30.13 & .9584 & .2228 & 32.16 & .9581 & .1717 & 28.94 & .9258 & .2635 & 31.23 & .9109 & .2578 & 27.15 & .9196 & .2622 & 29.92 & .9346 & .2356 \\   
        E\textsuperscript{3}NeRF   & 31.50 & \textbf{.9636} & \textbf{.2213} & \textbf{34.02} & \textbf{.9641} & \textbf{.1505} & \textbf{30.25} & \textbf{.9339} & \textbf{.2479} & \textbf{32.48} & \textbf{.9342} & \textbf{.1736} & \textbf{28.75} & \textbf{.9362} & \textbf{.2067} & \textbf{31.40} & \textbf{.9464} & \textbf{.2000} \\
    
    \bottomrule
    
    \end{tabular}
        \begin{tablenotes}
        \footnotesize
        \color{black}
        \item  \textbf{Ev-DeblurNeRF*}: Training Ev-DeblurNeRF with the poses estimated by our event-guided pose estimation model in Sec.~\ref{sec:4.3}.
    \end{tablenotes}
\end{table*}

\begin{figure*}[t]
    \centering
    \includegraphics[width=1\linewidth]{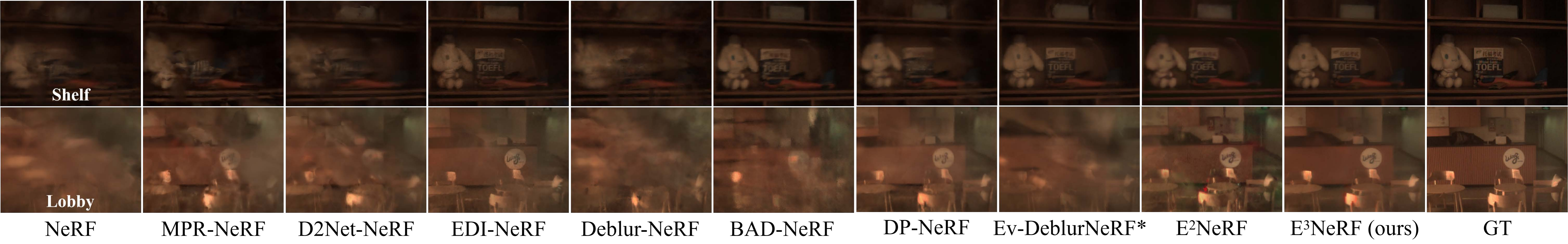}
    \caption{Qualitative results on ``Shelf'' and ``Lobby'' scenes of Real-World-Challenge dataset. \textcolor{black}{Ev-DeblurNeRF* is defined in TABLE~\ref{tab:5}.}}
    \label{fig:11}
\end{figure*}

\subsection{Results of Real-World-Blur Dataset}
\label{sec:5.5}
As in the conference paper of our E\textsuperscript{2}NeRF \cite{e2nerf}, we first compare our methods with NeRF, Deblur-NeRF, BAD-NeRF, D2Net-NeRF, MPR-NeRF, and EDI-NeRF on the Real-World-Blur dataset.
Since COLMAP misses some poses of the input images in the Real-World-Blur dataset, as shown in TABLE~\ref{tab:2}, we use the poses obtained by our pose estimation framework for all the compared methods to ensure they have fundamental results.

\myPara{Quantitative Results:}
We conduct quantitative analysis with two no-reference image quality assessment metrics, RankIQA\cite{rankiqa} and MetaIQA\cite{metaiqa}, because the Real-World-Blur dataset does not have ground-truth sharp images.
As shown in TABLE~\ref{tab:4}, 
Deblur-NeRF and BAD-NeRF present competitive performance on this forward-facing dataset compared to D2Net-NeRF and MPR-NeRF, but they still fall short of EDI-NeRF.
Our E\textsuperscript{2}NeRF and E\textsuperscript{3}NeRF achieve the second-best and best results, demonstrating the effectiveness of our models using events to enhance sharp neural radiance field reconstruction from blurry images.

\myPara{Qualitative Results:}
As shown in Fig.~\ref{fig:10}, MPR-NeRF fails to reconstruct a sharp NeRF.
The results of Deblur-NeRF and BAD-NeRF are superior to those of MPR-NeRF, but they still contain some blur and a significant amount of granular material.
With event data, D2Net-NeRF has a limited deblurring effect, and EDI-NeRF achieves a better deblurring effect.
However, a large number of noise events in low-light environments cause noisy results for EDI-NeRF.
Besides, EDI-NeRF misses the texture details on the grain of the lobster's back in the ``Toys'' scene, while our E\textsuperscript{3}NeRF effectively recovers these details and realizes the noiseless sharpest results compared to other methods.

\subsection{Results of Real-World-Challenge Dataset}
\label{sec:5.6}
To conduct a more comprehensive analysis, we compare our methods with MPR-NeRF, D2Net-NeRF, EDI-NeRF, Deblur-NeRF, BAD-NeRF, DP-NeRF, and Ev-DeblurNeRF on our Real-World-Challenge dataset, which features ground-truth sharp images for evaluation and more challenging input with severe blur.
Since COLMAP fails to estimate all poses of the input images in the Real-World-Blur dataset, as shown in TABLE~\ref{tab:2}, we use the poses obtained by our pose estimation framework for all the compared methods.
\textcolor{black}{We also use the extra poses of virtual sharp images of each view that are obtained by our framework as the reference poses for Ev-DeblurNeRF to implement pose interpolation for events, which ensures its fundamental performance.}

\myPara{Quantitative Results:}
We use PSNR, SSIM, and LPIPS \cite{lpips} to quantitatively evaluate the results of the Real-World-Challenge dataset.
As shown in TABLE~\ref{tab:5}, E\textsuperscript{3}NeRF achieves the best results on this dataset.
With the help of the event-guided spatial-temporal blur model, E\textsuperscript{3}NeRF is further improved compared to E\textsuperscript{2}NeRF.
EDI-NeRF outperforms MPR-NeRF and D2Net-NeRF, but still lags behind our E\textsuperscript{2}NeRF and E\textsuperscript{3}NeRF, which is consistent with the results in the synthetic data.
\textcolor{black}{
Deblur-NeRF and BAD-NeRF exhibit some basic effects in this dataset, but they are inferior to DP-NeRF.
However, these three methods are only effective for the forward-facing data.
Consequently, their performance drops significantly in ``lobby'' and ``table'' scenes, of which the input view distributions are close to a hemispherical shape.
Ev-DeblurNeRF is built upon DP-NeRF and also performs poorly in these two scenes.
In the other three scenes, Ev-DeblurNeRF achieves competitive performance and is better than our E\textsuperscript{2}NeRF on the ``shelf'' scene, but is still worse than our E\textsuperscript{3}NeRF because of lacking ground-truth poses and extra events for training.}
On the contrary, E\textsuperscript{3}NeRF ensures a stable pose estimation under extreme motion conditions and guides training to the spatial-temporal areas where blur appears.
Eventually, by explicitly simulating the blurring process of images and the generating process of events, a sharp NeRF is reconstructed from sparse and blurry input.

\myPara{Qualitative Results:}
As shown in Fig.~\ref{fig:11}, the blur in this dataset is much more severe than the Real-World-Blur dataset, causing a lot of cloud-like floating materials in the rendering results of the compared methods, which indicates that a wrong 3D representation is learned, especially for the ``Lobby'' scene.
\textcolor{black}{
BAD-NeRF and DP-NeRF achieve high-quality results on the ``Shelf'' scene, but the letters on the book are still not as sharp as our results.
Ev-DeblurNeRF performs better than DP-NeRF and E\textsuperscript{2}NeRF, but when it comes to the ``Lobby'' scene with spherical input view distribution, its reconstruction results will be significantly worse.
Our E\textsuperscript{3}NeRF shows strong robustness on the ``Lobby'' scene and has the best results in these two scenes.
}

\begin{table}[t]
    \caption{Quantitative results on TUM-VIE dataset for the comparison with event-based NeRF works and E\textsuperscript{3}NeRF.}
    \label{tab:6}
    \centering
    \setlength{\tabcolsep}{0.5mm}
    \begin{tabular}{c||cc||ccc|ccc}
        \toprule
        & \multicolumn{2}{||c||}{\textcolor{black}{Input Data}} & \multicolumn{3}{c|}{mocap-1d-trans}    & \multicolumn{3}{c}{mocap-desk2} \\
        
        \midrule
        & \textcolor{black}{Image} & \textcolor{black}{Event} & PSNR\textuparrow & SSIM\textuparrow & LPIPS\textdownarrow & PSNR\textuparrow & SSIM\textuparrow & LPIPS\textdownarrow \\
        \midrule
        E-NeRF                      & \textcolor{black}{-} & \textcolor{black}{\checkmark} & 15.32 & .6674 & .4614 & 12.57 & .7187 & .3333 \\ 
        \emph{e}-NeRF               & \textcolor{black}{-} & \textcolor{black}{\checkmark} & 15.60 & .7885 & .3008 & 16.17 & .8042 & .2602  \\
        \midrule
        
        E\textsuperscript{3}NeRF    & \textcolor{black}{\checkmark} & \textcolor{black}{\checkmark} & \textbf{30.18} & \textbf{.9606} & \textbf{.1512} & \textbf{26.49} & \textbf{.9470} & \textbf{.1706} \\
        \bottomrule
    \end{tabular}
\end{table}

\begin{figure}[t]
\centering
    \includegraphics[width=1\linewidth]{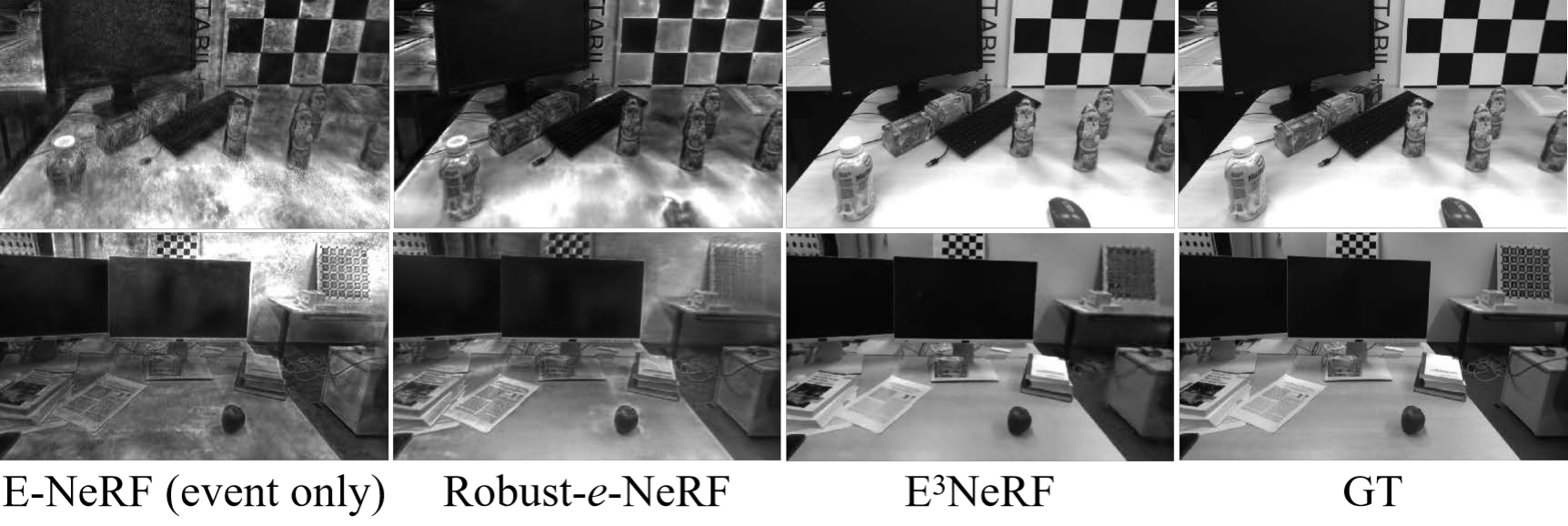}
    \caption{Qualitative results of mocap-1d-trans scene (first row) and mocap-desk2 scene (second row) on TUM-VIE dataset.}
    \label{fig:12}
\end{figure}

\subsection{Results of TUM-VIE Dataset}
\label{sec:5.7}
\textcolor{black}{
We also compare E\textsuperscript{3}NeRF with the event-based NeRF methods E-NeRF \cite{enerf2} and \emph{e}-NeRF \cite{renerf} to demonstrate the positive impact of introducing RGB data in event-based NeRF reconstruction.}
We use the ``mocap-1d-trans'' and ``mocap-desk2'' scenes of the TUM-VIE dataset to conduct the experiment as in their papers.
Note that the input views of E-NeRF and \emph{e}-NeRF are event-related and dense, whereas the input views of our method are image-related and sparse, consistent with our setting.
\textcolor{black}{
Since E-NeRF and \emph{e}-NeRF do not report the quantitative results in their papers, we retrained the models using their publicly available code and parameters \cite{enerf2-code, renerf-code} for the TUM-VIE dataset.
The results are shown in TABLE~\ref{tab:6} and Fig.~\ref{fig:12}.
}

\myPara{Quantitative Results:}
As shown in TABLE~\ref {tab:6}, event-based NeRF methods are challenging to achieve satisfactory results with only events for training, even when the events are captured from continuous, dense views.
Using both events and images as supervision, E\textsuperscript{3}NeRF performs significantly better with sparse view input.
This proves that reconstructing neural \textcolor{black}{radiance} fields using ERGB data is a more practical solution in low-light, high-speed scenes.

\myPara{Qualitative Results:}
The qualitative results in Fig.~\ref{fig:12} are consistent with the quantitative results.
The design for uncertain event camera parameters significantly improves the results of \emph{e}-NeRF compared to E-NeRF.
However, its reconstruction results are still not comparable to our method, even with more events, views, and ground-truth poses.

\begin{table*}[h]
    \caption{Quantitative results on the Ev-DeblurCDAVIS dataset for the comparison between ERGB-based NeRF methods and our E\textsuperscript{3}NeRF. The results are the averages of five scenes of the dataset.}
    \label{tab:7}
    \centering
    \setlength{\tabcolsep}{0.5mm}
    \begin{tabular}{c||cc||cccc|cccc|cccc}
        \toprule
        & & & \multicolumn{4}{c|}{Full Image Views as Input}    & \multicolumn{4}{c|}{1/2 Image Views as Input} & \multicolumn{4}{c}{1/3 Image Views as Input} \\
        
        & \textcolor{black}{GT Poses} & \textcolor{black}{Extra Events} & \textcolor{black}{Ev-Num} & PSNR\textuparrow & SSIM\textuparrow & LPIPS\textdownarrow & \textcolor{black}{Ev-Num} & PSNR\textuparrow & SSIM\textuparrow & LPIPS\textdownarrow & \textcolor{black}{Ev-Num} & PSNR\textuparrow & SSIM\textuparrow & LPIPS\textdownarrow \\
        \midrule
        E-NeRF*                    & \textcolor{black}{\checkmark} & \textcolor{black}{\checkmark} & \textcolor{black}{4787985} & 28.18 & 0.75 & 0.23 & - & - & - & - & - & - & - & - \\
        \midrule
        \multirow{3}*{Ev-DeblurNeRF}          
                                  & \textcolor{black}{\checkmark} & \textcolor{black}{\checkmark} & \textcolor{black}{4787985} & \textcolor{black}{32.96} & \textcolor{black}{0.91} & \textcolor{black}{0.06} & \textcolor{black}{4787985} & \textcolor{black}{32.75} & \textcolor{black}{0.95} & \textcolor{black}{0.06} & \textcolor{black}{4787985} & \textcolor{black}{32.16} & \textcolor{black}{0.95} & \textcolor{black}{0.06} \\
                                  & \textcolor{black}{\checkmark} & \textcolor{black}{-} & \textcolor{black}{4079610} & \textcolor{black}{32.63} & \textcolor{black}{0.95} & \textcolor{black}{0.06} & \textcolor{black}{2113999} & \textcolor{black}{28.82} & \textcolor{black}{0.90} & \textcolor{black}{0.15} & \textcolor{black}{1386928} & \textcolor{black}{24.92} & \textcolor{black}{0.81} & \textcolor{black}{0.29} \\
                                  & \textcolor{black}{-} & \textcolor{black}{\checkmark} & \textcolor{black}{4787985} & 29.47 & 0.91 & 0.07 & \textcolor{black}{4787985} & 28.56 & 0.90 & 0.08 & \textcolor{black}{4787985} & 28.51 & 0.90 & 0.09  \\

        \midrule
        Ev-DeblurNeRF             & \textcolor{black}{-} & \textcolor{black}{-} & \textcolor{black}{4079610} & 29.08 & 0.90 & \textbf{0.08} & \textcolor{black}{2113999} & 26.82 & 0.85 & 0.17 & \textcolor{black}{1386928} & 24.97 & 0.80 & 0.26  \\
        E\textsuperscript{3}NeRF  & \textcolor{black}{-} & \textcolor{black}{-} & \textcolor{black}{4079610} & \textbf{31.34} & \textbf{0.94} & \textbf{0.08} & \textcolor{black}{2113999} & \textbf{31.13} & \textbf{0.94} & \textbf{0.09} & \textcolor{black}{1386928} & \textbf{30.08} & \textbf{0.92} & \textbf{0.10} \\
        \bottomrule
    \end{tabular}
    
    \begin{tablenotes}
        \footnotesize
        \color{black}
        \item \textbf{E-NeRF*}: Training E-NeRF \cite{enerf2} with both events and RGB images, and the results are reported in \cite{evdeblurnerf}.
        \item \textbf{GT Poses}: The ground-truth camera poses provided by the dataset.
        \item \textbf{Extra Events}: The continuous dense view event streams that connect the adjacent sparse views of RGB images.
        \item  \textbf{Ev-Num}: The average number of events used for training of the five scenes.
    \end{tablenotes}
  
\end{table*}

\begin{figure*}[h]
    \centering
    \includegraphics[width=1\linewidth]{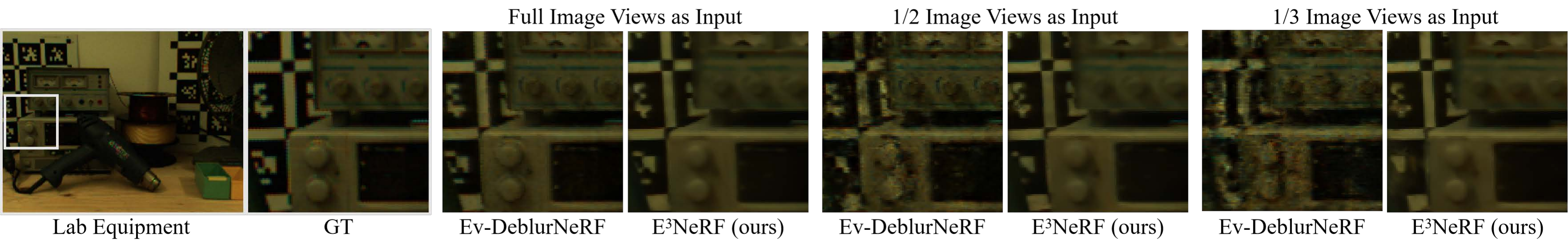}
    \color{black}
    \caption{Qualitative results on the ``Lab Equipment'' scene of the Ev-DeblurCDAVIS dataset without ground-truth poses and extra events.}
    \label{fig:13}
\end{figure*}

\subsection{Results on Ev-DeblurCDAVIS Dataset}
\label{sec:5.8}
We also conduct an additional experiment on the Ev-DeblurCDAVIS \cite{evdeblurnerf} dataset that provides both ground-truth poses and extra events for E-NeRF \cite{enerf2} and Ev-DeblurNeRF, and further analyze the advantages of our method in practical applications with sparse input compared to existing ERGB-based NeRF methods.

\myPara{Quantitative Results:}
\textcolor{black}{
As shown in the left part of TABLE~\ref{tab:7}, when there are full image views as input, E\textsuperscript{3}NeRF is better than Ev-DeblurNeRF under our settings without extra events and ground-truth poses.
Ev-DeblurNeRF with extra events as input exhibits better performance, but remains inferior to E\textsuperscript{3}NeRF in terms of PSNR and SSIM.
With both ground-truth poses and extra events as input, E-NeRF* is still worse than E\textsuperscript{3}NeRF, and Ev-DeblurNeRF shows better results than E\textsuperscript{3}NeRF.
However, such ground-truth poses are always difficult to obtain in practical applications.
}

\myPara{Robustness to the Sparse Input:}
\textcolor{black}{
Since the original goal of NeRF is to perform novel view synthesis from sparse view inputs, we reduce the input image views of the Ev-DeblurCDAVIS dataset to further analyze the robustness to sparse input of E\textsuperscript{3}NeRF and Ev-DeblurNeRF.
As in the right part of TABLE~\ref{tab:7}, when there are no extra events, the performance of Ev-DeblurNeRF drops significantly as the number of input images and events decreases, and when the extra events are provided, its performance decline is somewhat alleviated but still worse than E\textsuperscript{3}NeRF.
Only with both ground-truth poses and extra events as input, Ev-DeblurNeRF outperforms our method, demonstrating that Ev-DeblurNeRF's performance is highly dependent on the ground-truth poses and extra events.
Therefore, our method shows better practicability and robustness than the existing ERGB-Based NeRF methods on the real scenes where ground-truth poses and extra events are often not provided.
}

\myPara{Qualitative Results:}
\textcolor{black}{
As shown in Fig.~\ref{fig:13}, our method outperforms Ev-DeblurNeRF under the same input, which is consistent with the quantitative comparison results.
For the sparse input, the results of Ev-DeblurNeRF exhibit noticeable artifacts as the input data decreases, whereas E\textsuperscript{3}NeRF does not show noticeable performance degradation, indicating that our method is more robust to sparse input and more aligned with the original goal of NeRF.
}

\myPara{Discussion and Analysis:}
\textcolor{black}{
We notice that the performance of Ev-DeblurNeRF without GT poses and extra events is relatively worse on the Real-World-Challenge dataset than on the Ev-DeblurCDAVIS dataset.
This is because the input image and event views of the Real-World-Challeng dataset are relatively sparse and randomly distributed around the scene as discussed in Sec.~\ref{sec:5.1.3} and Fig.~\ref{fig:7}, which is unfavorable for the Ev-DeblurNeRF's pose interpolation strategy to calculate the precise pose for each event with given poses for reference. 
On the contrary, Ev-DeblurCDAVIS dataset is captured by a camera moving on a fixed straight track, meaning that the input image and event views are evenly arranged along a straight line in chronological order, which makes it possible to use linear pose interpolation to calculate the precise pose of each event with ground-truth poses for reference.
This further demonstrates that the E\textsuperscript{3}NeRF and Ev-DeblurNeRF models can achieve the best performance in their respective settings.
In contrast, our settings are more inclined to follow the application of novel view synthesis in the wild scenes, and the settings of Ev-DeblurNeRF are more inclined to the application of event-based 3D represent learning under an ideal environment in the laboratory.
}

\begin{table*}[!t]
    \caption{Ablation study on severe shake synthetic data and Real-World-Challenge dataset. The results are averages of all scenes and all test views.}
    \label{tab:8}
    \centering
    \setlength{\tabcolsep}{1mm}
    \begin{tabular}{l|ccccc||cccc|cccc}
        \toprule
        & \multirow{2}*{$\mathcal{L}_{blur}$} & \multirow{2}*{$\mathcal{L}_{event}$} & Temporal Blur & Spatial Blur & Motion-Guided & \multicolumn{4}{c|}{Synthetic Data} & \multicolumn{4}{c}{Real-World-Challenge Dataset}\\
        & & & Uniform Binning & Locating & Event Splitting & PSNR\textuparrow & SSIM\textuparrow & LPIPS\textdownarrow & Time & PSNR\textuparrow & SSIM\textuparrow & LPIPS\textdownarrow & Time \\
        \midrule
        NeRF                        & - & - & - & - & - &                                       21.02 & .8881 & .2094 & \textbf{6.0 h}  & 26.25 & .8864 & .4515 & \textbf{0.63 h} \\
        
        E\textsuperscript{2}NeRF*   & \checkmark & - & - & - & - 
        & 23.85 & .9179 & .1408 & 19.8 h & 28.79 & .9201 & .2939 & 2.85 h \\ 
        E\textsuperscript{2}NeRF    & \checkmark & \checkmark & - & - & - 
        & 26.75 & .9512 & .0924 & 20.1 h & 29.92 & .9346 & .2356 & 3.10 h \\
        E\textsuperscript{3}NeRF**  & \checkmark & \checkmark & \checkmark & - & - 
        & 27.96 & .9573 & .0685 & 20.5 h & 30.91 & .9432 & .2127 & 3.13 h \\ 
        E\textsuperscript{3}NeRF*   & \checkmark & \checkmark & \checkmark & \checkmark & - 
        & 28.97 & .9605 & .0606 & 10.3 h & 31.19 & .9440 & .2025 & 2.00 h \\ 
        E\textsuperscript{3}NeRF    & \checkmark & \checkmark & \checkmark & \checkmark & \checkmark 
        & \textbf{29.05} & \textbf{.9623} & \textbf{.0552} & 12.0 h & \textbf{31.40} & \textbf{.9464} & \textbf{.2000} & 2.87 h \\ 
        \bottomrule
    \end{tabular}
\end{table*}

\begin{figure*}[!t]
    \centering
    \includegraphics[width=1\linewidth]{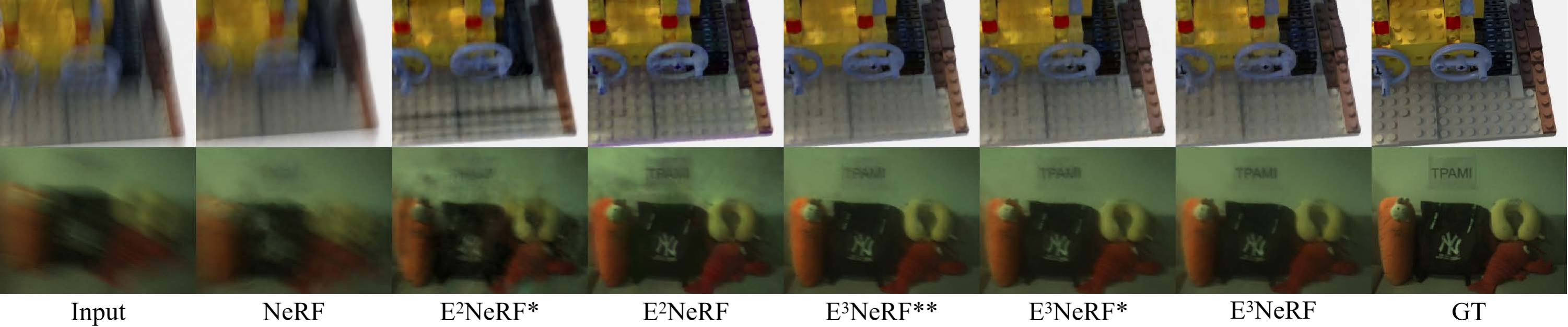}
    \caption{Ablation study on synthetic ``Lego'' and real-world ``Lab'' scenes. E\textsuperscript{2}NeRF*, E\textsuperscript{3}NeRF**, and E\textsuperscript{3}NeRF* are defined in TABLE~\ref{tab:8}}.
    \label{fig:14}
\end{figure*}

\subsection{Ablation Study and Analysis}
\label{sec:5.9}

\subsubsection{Effectiveness of Blur Loss and Event Loss}
\label{sec:5.9.1}
\myPara{Blur Loss:}
The results of NeRF and E\textsuperscript{2}NeRF* in TABLE~\ref{tab:8} and Fig.~\ref{fig:14} demonstrate that the blur rendering loss provides a substantial performance boost to our model.
A similar blur loss is also used in Deblur-NeRF and BAD-NeRF.
However, experiments in Secs.~\ref{sec:5.4}, Sec.~\ref{sec:5.5}, and Sec.~\ref{sec:5.6} show that relying solely on blur loss cannot achieve precise synthesis of the blurring process and high-fidelity reconstruction of the sharp NeRF from blurry input.

\begin{figure}[h]
    \centering
    \includegraphics[width=1\linewidth]{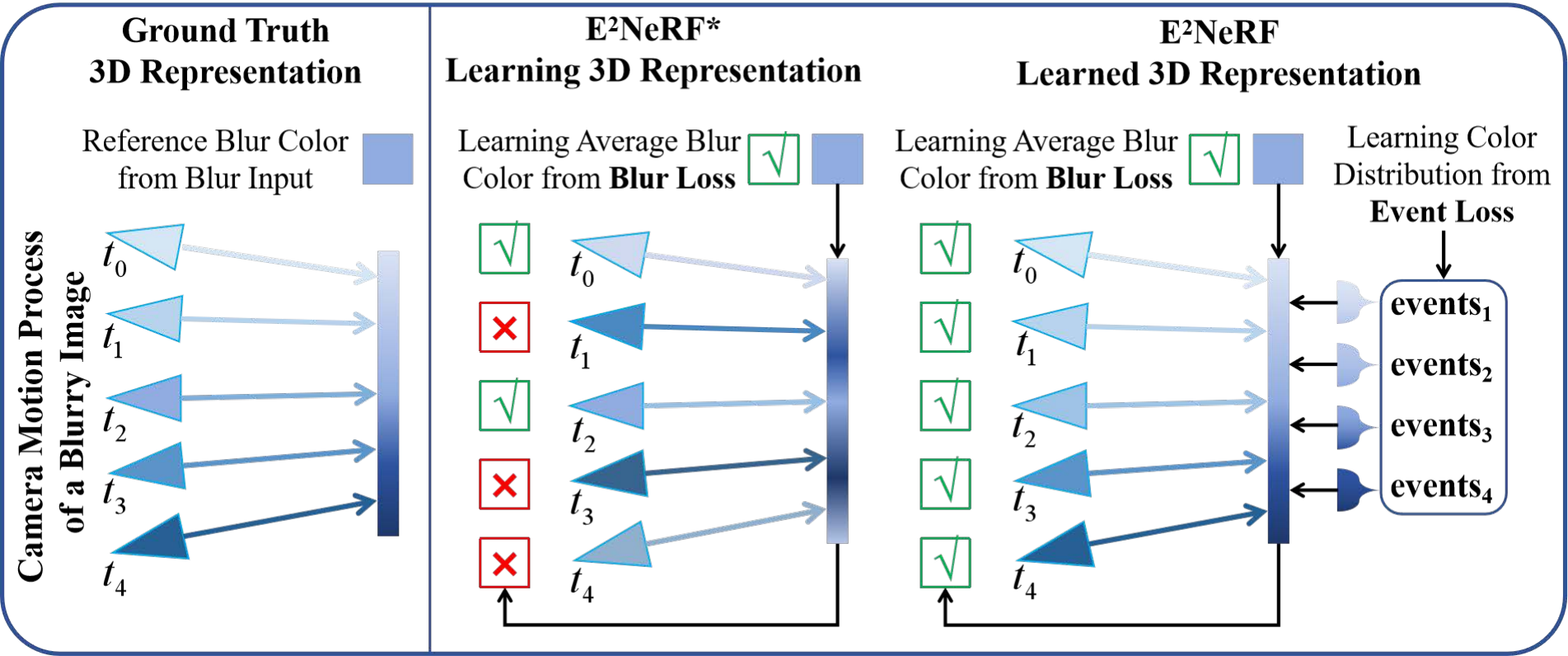}
    \caption{Effectiveness of event loss. E\textsuperscript{2}NeRF* denotes training E\textsuperscript{2}NeRF without event loss as supervision. As shown in the figure, E\textsuperscript{2}NeRF* and E\textsuperscript{2}NeRF both obtain the correct blur color with blur loss. However, without event loss supervising the light intensity change, E\textsuperscript{2}NeRF* tends to learn an inaccurate 3D representation. When supervised additionally with event loss, E\textsuperscript{2}NeRF can sort out the correct spatial color distribution and obtain results closer to the ground truth.}
    \label{fig:15}
\end{figure}

\myPara{Event Loss:}
The results of E\textsuperscript{2}NeRF* and E\textsuperscript{2}NeRF in TABLE~\ref{tab:8} show that the event loss further improves the performance.
We further explain this in Fig.~\ref{fig:15}.
As shown in the middle of the figure, the model learns the correct average color of the scene with blur loss; however, the color distribution at different poses remains uncertain.
Therefore, there will be situations where the mean value is correct but the individual values are wrong.
This is reflected in the alternating light and dark lines along the motion blur direction as the blue line in the middle of Fig.~\ref{fig:15}.
The abnormal horizontal stripes in the results of E\textsuperscript{2}NeRF* on the ``Lego'' scene in Fig.~\ref{fig:14} are exactly related to this.
With event loss as supervision of the brightness change information between $t_{k}$ and $t_{k+1}$, E\textsuperscript{2}NeRF can accurately associate each pose with the rendered color and eliminate this uncertainty, ultimately learning an accurate neural 3D representation, which yields better novel view rendering results as shown in Fig.~\ref{fig:14}.

\subsubsection{Effectiveness of Spatial-Temporal Blur Model}
\label{sec:5.9.2}
\myPara{Event-Guided Temporal Blur Uniform Binning:}
Temporal blur uniform binning distributes training evenly over the temporal blur and has a significant improvement over E\textsuperscript{2}NeRF as shown in the fourth row of TABLE~\ref{tab:8}.
The cloudy materials are also eliminated in the results of E\textsuperscript{3}NeRF** as in the second row of Fig.~\ref{fig:14}.
Notice that the performance improvement is more evident in real-world data, proving the practicality of the model.

\begin{table}[h]
    \color{black}
    \caption{A detailed analysis of spatial blur locating on the Real-World-Challenge dataset. Results are the averages of the five scenes.}
    \label{tab:9}
    \centering
    \setlength{\tabcolsep}{1.8mm}
    \begin{tabular}{c||cc|ccc}
        \toprule
        & \multicolumn{2}{c|}{Spatial Blur Locating} & \multicolumn{3}{c}{Real-World-Challenge} \\
        & $\mathcal{L}_{blur}$ & $\mathcal{L}_{event}$ & PSNR\textuparrow & SSIM\textuparrow & LPIPS\textdownarrow \\

        \midrule
        \multirow{2}*{E\textsuperscript{3}NeRF**\textsubscript{nob}}
        & \multirow{2}*{N/A} & - & 29.23 & .9341 & .2735 \\
        & & \checkmark & 30.33 & .9407 & .2234 \\
        
        \midrule
        \multirow{2}*{E\textsuperscript{3}NeRF**\textsubscript{noe}}
        & - & \multirow{2}*{N/A} & 28.79 & .9201 & .2939 \\
        & \checkmark & & 27.07 & .8990 & .4081 \\
        
        \midrule
        \multirow{3}*{E\textsuperscript{3}NeRF**}
        & - & - & 30.91 & .9432 & .2127 \\ 
        & \checkmark & - & 30.24 & .9406 & .2311 \\
        & - & \checkmark & 30.90 & .9433 & .2073 \\

        \midrule
        E\textsuperscript{3}NeRF* & \checkmark & \checkmark & 31.19 & .9464 & .2025 \\
        \bottomrule
    \end{tabular}

    \begin{tablenotes}
        \footnotesize
        \color{black}
        \item \textbf{\checkmark:} Implementing the spatial blur locating on the loss.
        \item \textbf{- :} Not implementing the spatial blur locating on the loss.
        \item \textbf{E\textsuperscript{3}NeRF**\textsubscript{nob}:} Training E\textsuperscript{3}NeRF** in TABLE~\ref{tab:8} without blur loss.
        \item \textbf{E\textsuperscript{3}NeRF**\textsubscript{noe}:} Training E\textsuperscript{3}NeRF** in TABLE~\ref{tab:8} without event loss.
        \item  \textbf{E\textsuperscript{3}NeRF*:} Training E\textsuperscript{3}NeRF without motion-guided event splitting. 
    \end{tablenotes}
\end{table}

\myPara{Event-Guided Spatial Blur Locating:}
Spatial blur locating focuses the training on blurry areas and reduces the training on smooth areas, improving our model's training efficiency by 49.8\% and 36.1\% over E\textsuperscript{2}NeRF on synthetic data and real data, respectively, as shown in the fifth row in TABLE~\ref{tab:8}.
The blurred areas in the synthetic data account for less, and therefore, the efficiency improvement is more prominent.
In comparison, Deblur-NeRF takes 19.6 and 2.7 hours in the synthetic and real data, respectively, under the same training conditions.
Additionally, the quantitative and qualitative results in TABLE~\ref{tab:8} and Fig.~\ref{fig:14} also show that spatial blur locating improves the results on both synthetic and real-world data.
\textcolor{black}{
We further analyze this phenomenon on the Real-World-Challenge dataset by applying spatial blur locating to blur loss and event loss separately for E\textsuperscript{3}NeRF**\textsubscript{nob}, E\textsuperscript{3}NeRF**\textsubscript{noe}, and E\textsuperscript{3}NeRF**.
The results are shown in TABLE~\ref{tab:9}.
Firstly, we limit event loss to the blur areas with the spatial blur locating for E\textsuperscript{3}NeRF**\textsubscript{nob}, and it produces better results because there are no valid events in the sharp areas to supervise event loss effectively.
In other words, the constraint of ${\rm sum}(B_{k}(\mathbf{x}))=0$ in Eq.~\eqref{eq:24} is too strict for the sharp areas without events, leading to a negative impact of event loss in sharp areas.
Secondly, although the results of E\textsuperscript{3}NeRF**\textsubscript{noe} show that limiting blur loss to the blur areas with spatial blur locating will have a negative impact, comparing the fifth row with the first and third rows of TABLE~\ref{tab:9}, E\textsuperscript{3}NeRF** without spatial blur locating demonstrate that blur loss and event loss are a pair of complementary supervisory signals during training, and only when they are applied simultaneously on the same area can the best effect be achieved.
Consequently, if we only apply spatial blur locating to event loss or blur loss individually for E\textsuperscript{3}NeRF**, the performance will be reduced due to the separation of the two loss functions' acting area, as shown in the sixth and seventh rows in TABLE~\ref{tab:9}.
Notice that the performance drop in the sixth row is not apparent because the performance improvement of applying event loss only to the blurred areas offsets the adverse effect of separating the two losses.
Therefore, E\textsuperscript{3}NeRF* applies spatial blur locating simultaneously on blur loss and event loss, which avoids the negative impact of event loss in the sharp areas without separating the two losses, achieving the best performance as shown in the last row of TABLE~\ref{tab:9}.
}

\subsubsection{Analysis on Motion-Guided Event Splitting}
\label{sec:5.9.3}
In this section, we first evaluate the influence of the fixed $b$ on E\textsuperscript{2}NeRF, and further analyze the importance of the proposed motion-guided event splitting in real-world applications with varying degrees of blur across different views.

\begin{figure}[h]
    \centering
    \includegraphics[width=1\linewidth]{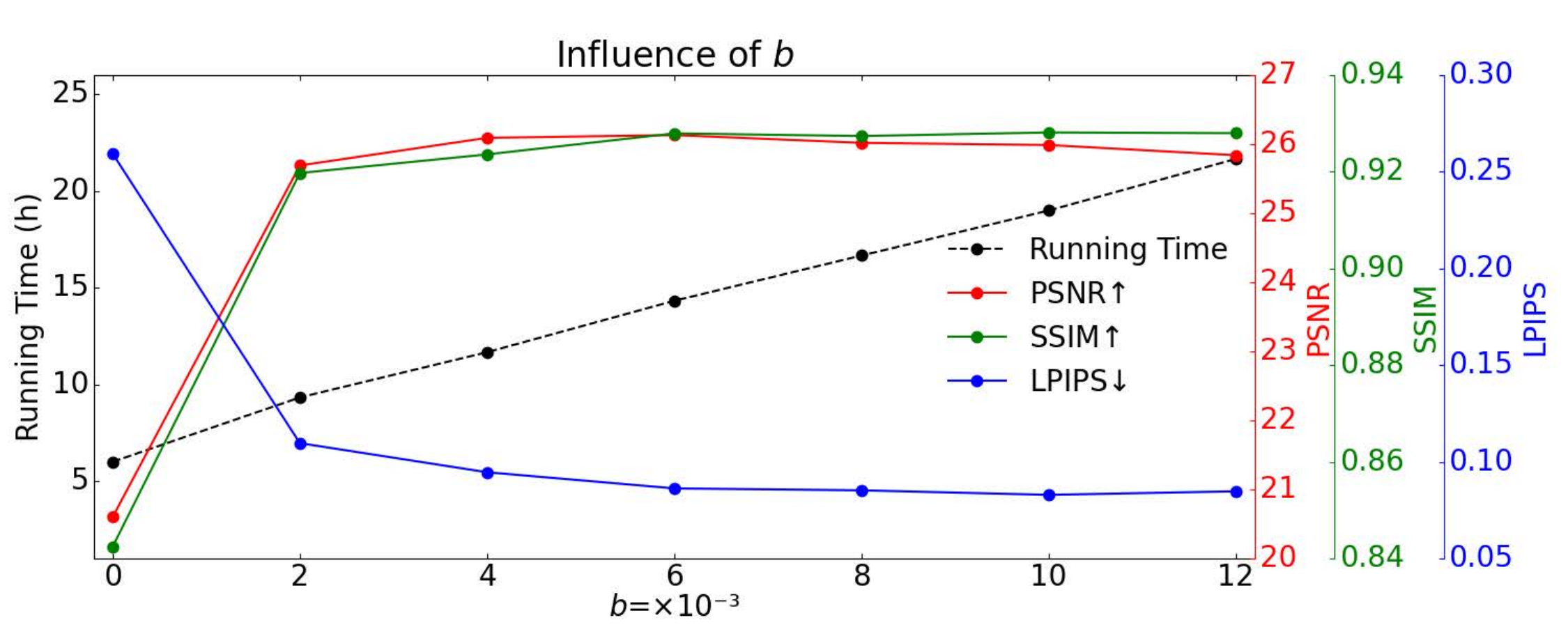}
    \caption{Analysis on influence of $b$ on synthetic ``Lego'' scene.}
    \label{fig:16}
\end{figure}

\myPara{Influence of $b$:}
E\textsuperscript{2}NeRF uses the same fixed $b$ for all input views.
As shown in Fig.~\ref{fig:16}, as $b$ increases from $b=0$ (original NeRF), the results gradually improve, but at the same time, the training time increases because the network needs to render more virtual sharp frames.
In E\textsuperscript{2}NeRF, we set $b=4$ as a trade-off between performance and training time, because when $b>4$ the results are barely improved.

\begin{figure}[h]
    \centering
    \includegraphics[width=1\linewidth]{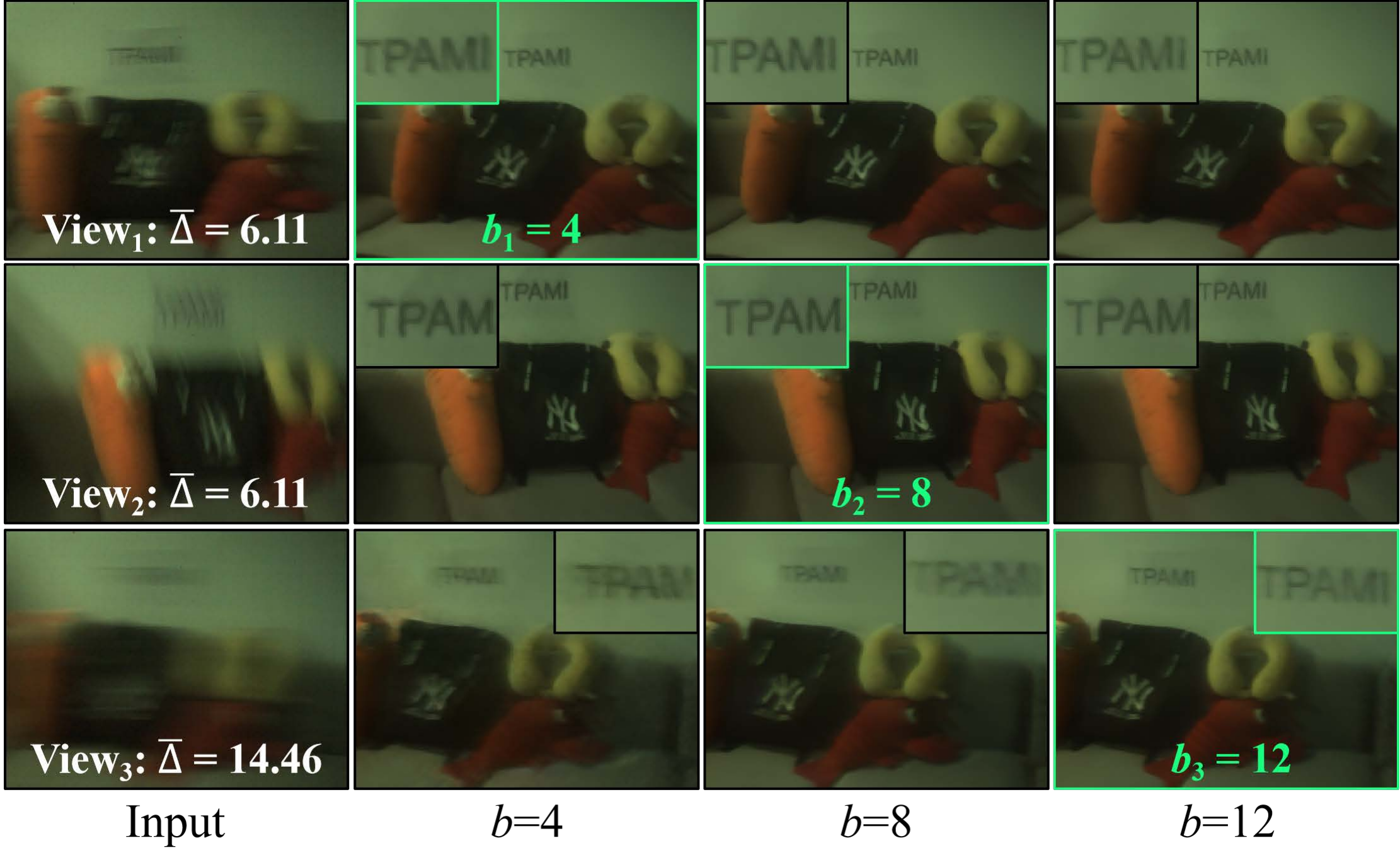}
    \caption{Analysis of the motion-guided event splitting on real-world ``Lab'' scene. More severely blurred input images often require a larger $b$ to achieve satisfactory results. The green boxes show the calculated $b_{v}$ for each view of our E\textsuperscript{3}NeRF with the motion-guided event splitting model.}
    \label{fig:17}
\end{figure}

\myPara{Motion-Guided Event Splitting:}
We show three input views of the ``Lab'' scene characterized by distinct motion speed in the left part of Fig.~\ref{fig:17}.
The corresponding blur ranges $\bar\Delta_{i}$ for the three views calculated by our motion-guided event-splitting model are 5.25, 6.11, and 14.46, respectively.
The results for different values of $b$ are shown in the right part of Fig.~\ref{fig:17}.
For a view with a slightly blurry input image, $b=4$ is sufficient to complete a sharp reconstruction, as shown in the first row.
However, for a view with more severe blurry input, as in the second row, some artifacts occur in the result of $b=4$.
For an extremely blurry input, as in the third row, the result is distorted even with $b=8$.
Our proposed motion-guided event splitting can adjust the appropriate $b_{v}$ for each view as shown in the green box of Fig.~\ref{fig:17}.
As a result, the performance of E\textsuperscript{3}NeRF is improved compared to E\textsuperscript{3}NeRF* without a significant increase in training time, especially for the Real-World-Challenge dataset, as in the last two rows of TABLE~\ref{tab:8}.

\subsubsection{Influence of the Loss Weight and Threshold}
\label{sec:5.9.4}

\begin{figure}[h]
    \centering
    \includegraphics[width=1\linewidth]{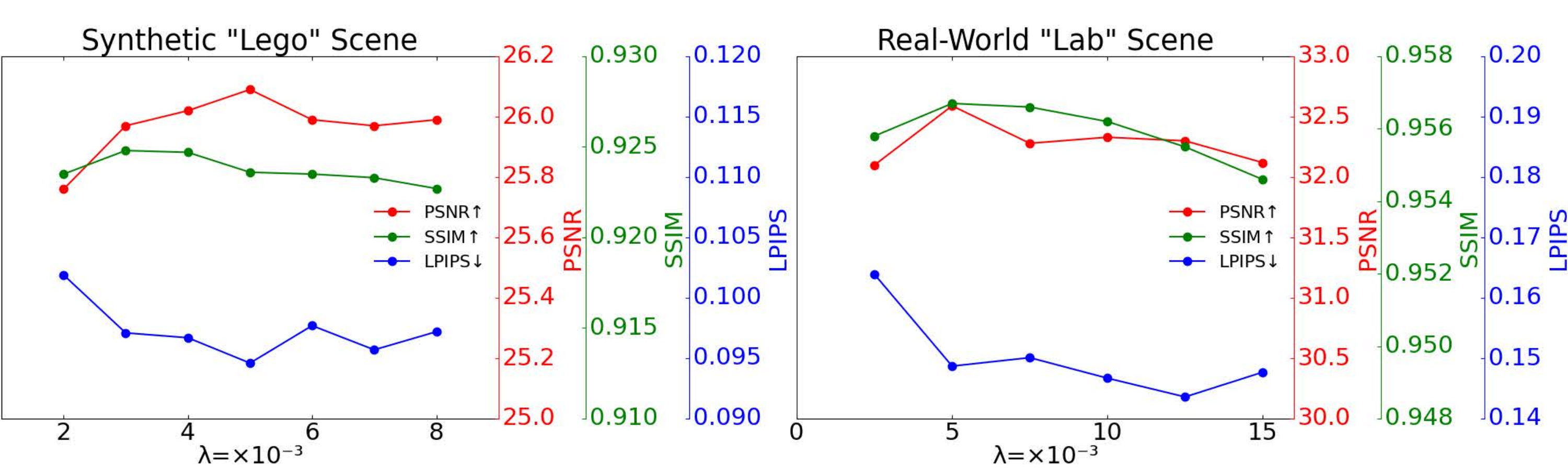}
    \caption{Analysis on the influence of event loss weight $\lambda$ on Synthetic ``Lego'' scene and real-world ``Lab'' scene.}
    \label{fig:18}
\end{figure}

\myPara{Loss Weight $\lambda$:}
Fig.~\ref{fig:18} demonstrates that for both synthetic and real-world data, the performance initially improves and then declines as $\lambda$ gradually increases from 0.001, and the overall optimal results are obtained when $\lambda=0.005$.

\begin{table}[t]
\caption{Analysis on influence of threshold $\Theta$ on real-world ``Lab'' scene.}
\label{tab:10}
    \centering
    \setlength{\tabcolsep}{1.5mm}
    \begin{tabular}{c||c|cccccc}
        \toprule
        & $\Theta$            & 0.1 & 0.2 & 0.3 & 0.4 & 0.5 & 0.6 \\
        \midrule
        \multirow{3}*{\shortstack{Real-World\\``Lab'' Scene}}
        & PSNR\textuparrow    & 26.70 & 30.73 & 34.02 & 33.78 & 33.55 & 32.22 \\ 
        & SSIM\textuparrow    & .8998 & .9418 & .9641 & .9655 & .9632 & .9521 \\
        & LPIPS\textdownarrow & .2775 & .1956 & .1505 & .1506 & .1487 & .1585 \\
        \bottomrule
    \end{tabular}
\end{table}

\myPara{Threshold $\Theta$:}
In TABLE~\ref{tab:10}, we train E\textsuperscript{3}NeRF with different thresholds of event loss on the real-world ``Lab'' scene.
When $\Theta=0.3$, the overall performance of the three metrics is the best, aligning with the default threshold settings of the event camera during data capture.

\subsection{Discussions and Limitations}
\label{sec:5.10}

\myPara{Compatibility with Recent 3D Represent Learning Models:}
Recently, some 3D representation learning models, such as Instant-NGP\cite{instantngp} and 3D Gaussian Splatting (3DGS)\cite{3dgs}, have made significant improvements in rendering and training speed compared to NeRF.
Some works have attempted to migrate our conference work E\textsuperscript{2}NeRF to 3DGS, such as E2GS\cite{e2gs}, EvaGaussians\cite{evags}, and EaDeblur-GS\cite{eadeblurgs}, demonstrating the generalization of our proposed blur loss, event loss, and pose estimation framework across different backbones.
Thanks to the powerful performance of 3DGS, their rendering and training speeds have also been greatly improved.
Besides, our proposed event-guided spatial-temporal blur model in E\textsuperscript{3}NeRF is also compatible with the 3DGS framework.
Thus, we believe our E\textsuperscript{3}NeRF is also backbone-agnostic and will significantly benefit the ERGB-based 3DGS.
Our future work will also further explore this in more challenging application scenes.

\myPara{Limitations:}
The event-guided pose estimation framework in E\textsuperscript{3}NeRF is still based on COLMAP.
Some image-based NeRF works have explored eliminating COLMAP for pose initialization, opting instead to learn the implicit 3D representation jointly with the camera poses.
For example, BARF\cite{barf} uses a coarse-to-fine strategy to gradually increase the position encoding dimension during training.
L2G-NeRF\cite{l2g-nerf} follows BARF and introduces a local-to-global module, achieving better results under large pose disturbance.
Nope-NeRF\cite{nope-nerf} imports depth information to supervise the joint optimization of NeRF and camera poses.
We believe that events can also play a crucial role in the joint optimization of poses and neural radiance fields, and a COLMAP-free deblurring NeRF utilizing both event and image data could represent a promising future direction for improvement.

\section{Conclusion}
\label{sec:6}
In this paper, we propose a novel \textbf{E}fficient \textbf{E}vent-\textbf{E}nhanced NeRF (E\textsuperscript{3}NeRF), the first framework for learning a sharp neural 3D representation from blurry images and corresponding event data.
Two novel losses are proposed to establish the connection between images, events, and neural radiance fields.
An event-guided spatial-temporal blur model, based on the correlation between image motion blur and event distribution, is introduced to unlock the potential of the network.
We demonstrate the effectiveness of the proposed model on both synthetic and real-world datasets.
The results indicate that our framework has significant improvement over other image-based deblurring NeRF, event-based NeRF, ERGB-based NeRF, and image-deblurring approaches.
Overall, we believe that our work will shed light on the research of high-quality 3D representation learning with ERGB data in complex and low-light scenes.

\section*{Acknowledgments}
This work is partially supported by the National Natural Science Foundation of China under Grant 62132002 and the Fundamental Research Funds for the Central University.

\bibliographystyle{IEEEtran}
\bibliography{egbib}

\begin{IEEEbiography}[{\includegraphics[width=1in,height=1.25in,clip,keepaspectratio]{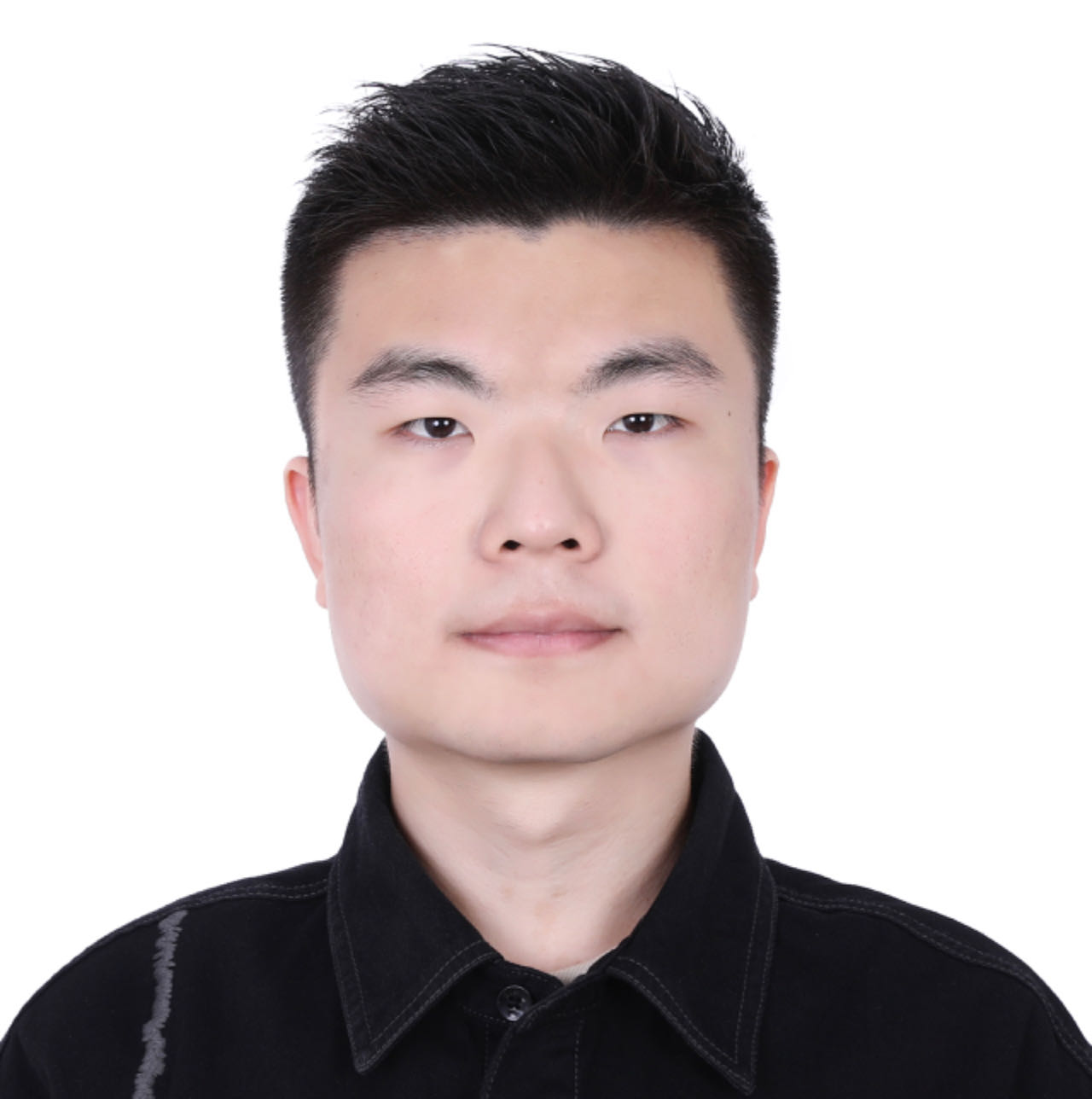}}]{Yunshan Qi} received the B.E. degree from Beihang University in 2020. He is currently working toward a Ph.D. degree with the State Key Laboratory of Virtual Reality Technology and Systems, School of Computer Science and Engineering, Beihang University, Beijing, China. His research interests include computer vision, event-based vision, and neural radiance fields.
\end{IEEEbiography}

\begin{IEEEbiography}[{\includegraphics[width=1in,height=1.25in,clip,keepaspectratio]{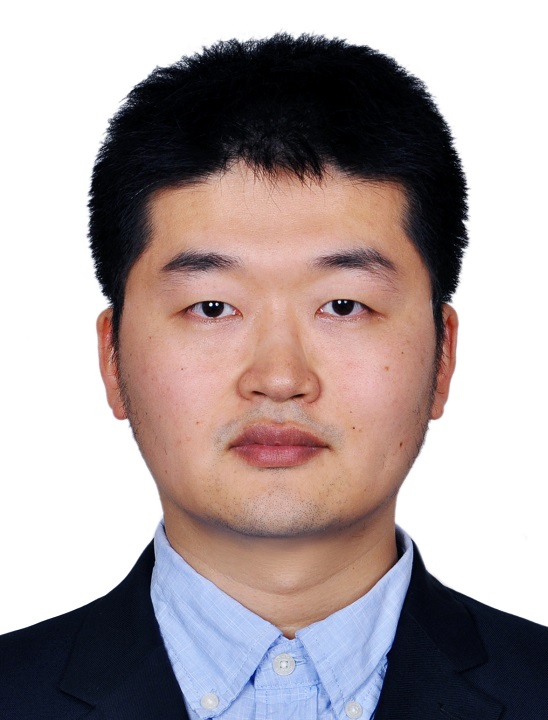}}] {Jia Li} is currently a Full Professor with the State Key Laboratory of Virtual Reality Technology and Systems, School of Computer Science and Engineering, Beihang University. He received his B.E. degree from Tsinghua University in 2005 and his Ph.D. degree from the Institute of Computing Technology, Chinese Academy of Sciences, in 2011. Before he joined Beihang University in 2014, he used to work at Nanyang Technological University, Shanda Innovations, and Peking University. His research is focused on computer vision, multimedia, and artificial intelligence, especially visual computing in extreme environments. He has co-authored more than 120 articles in peer-reviewed top-tier journals and conferences. He also has one Monograph published by Springer and more than 70 patents issued in the U.S. and China. He is a Fellow of IET, a Distinguished Member of CCF, and a Senior Member of IEEE/ACM/CCF/CIE.
\end{IEEEbiography}

\begin{IEEEbiography}[{\includegraphics[width=1in,height=1.25in,clip,keepaspectratio]{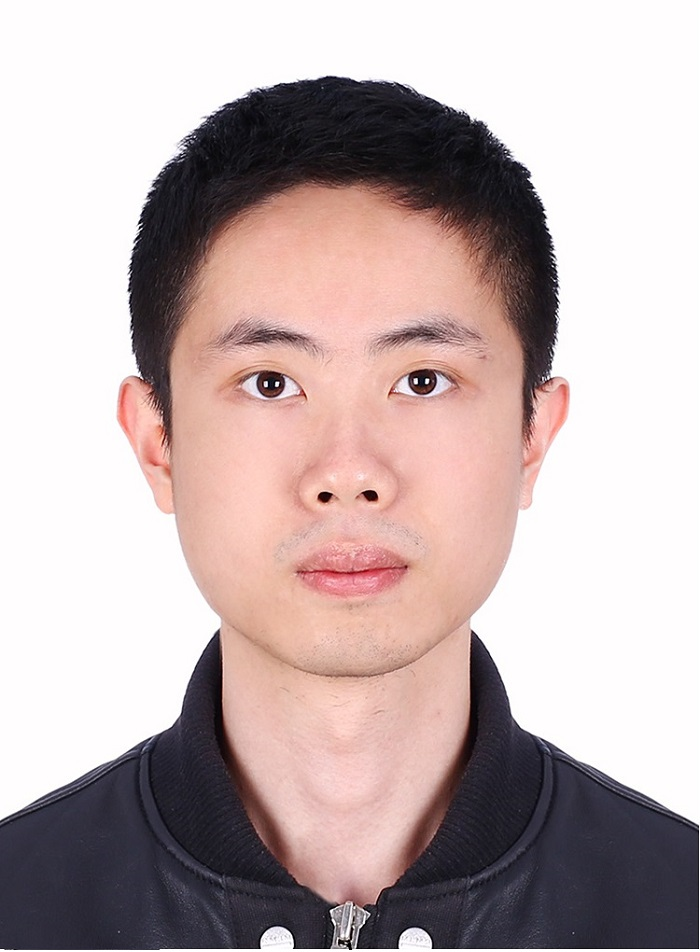}}] {Yifan Zhao} (Member, IEEE) is currently an Associate Professor with the School of Computer Science and Engineering, Beihang University, Beijing, China. He worked as a Boya Postdoc researcher with the School of Computer Science, Peking University. He received the B.E. degree from Harbin Institute of Technology in Jul. 2016 and the Ph.D. degree from the School of Computer Science and Engineering, Beihang University, in Oct. 2021. His research interests include computer vision and image/video understanding.
\end{IEEEbiography}

\begin{IEEEbiography}[{\includegraphics[width=1in,height=1.25in,clip,keepaspectratio]{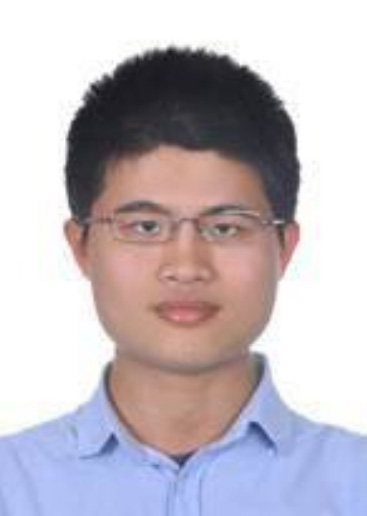}}] {Yu Zhang} is currently a researcher with SenseTime Research, Beijing, China. He received his B.E. degree in 2012 and Ph.D. degree in 2018 from the School of Computer Science and Engineering, Beihang University. His research interests include computer vision and image/video processing.
\end{IEEEbiography}

\begin{IEEEbiography}[{\includegraphics[width=1in,height=1.25in,clip,keepaspectratio]{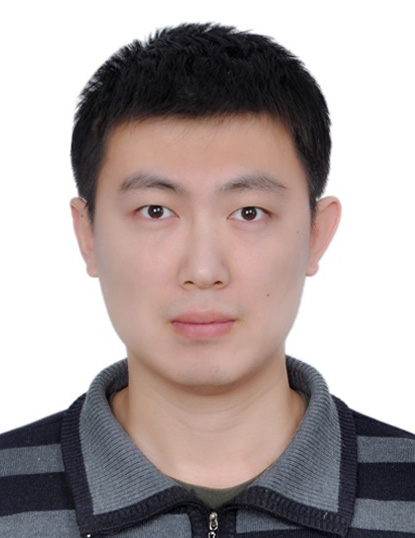}}] {Lin Zhu} received the B.S. degree in computer science from the Northwestern Polytechnical University, China, in 2014, the M.S. degree in computer science from the North Automatic Control Technology Institute, China, in 2018, and the Ph.D. degree with the School of Electronics Engineering and Computer Science, Peking University, China, in 2022. He is currently an assistant professor at the School of Computer Science, Beijing Institute of Technology, China. His current research interests include neuromorphic computing and event-based vision.
\end{IEEEbiography}

\end{document}